\documentclass[journal,letterpaper]{IEEEtran}
\usepackage{cite}
\usepackage{textcomp}
\usepackage[numbers]{natbib}
\usepackage{booktabs}
\usepackage{moreverb}
\usepackage{algorithmic}
\usepackage{graphicx}
\usepackage{url}
\usepackage{multicol,lipsum}
\usepackage{mathtools}
\usepackage{cuted}
\usepackage{multirow}
\usepackage{bm}
\usepackage{color, colortbl}
\usepackage[perpage]{footmisc}
\usepackage[colorlinks,bookmarksopen,bookmarksnumbered,citecolor=red,urlcolor=red, bookmarks=false]{hyperref} 
\usepackage{enumitem}
\usepackage{amsmath,amssymb,amsfonts}

\hypersetup
{
	pdftitle = {Model Predictive Control with Environment Adaptation for Legged Locomotion},
	pdfauthor = {Niraj Rathod},
	pdfsubject = {IEEE Access manuscript},
	pdfkeywords = {Legged locomotion, Mobility, Nonlinear Model Predictive Control, Online re-planning},
	pdftoolbar = true,
	colorlinks = true,
	linkcolor = black,
	citecolor = black,
	urlcolor = black,
}
\usepackage[usenames,dvipsnames]{xcolor}
\definecolor{blue_iit}{RGB}{51,51,255}
\usepackage{glossaries}
\usepackage[tight]{units}
\usepackage[normalem]{ulem} 
\usepackage{cancel}
\definecolor{Gray}{gray}{0.9}
\definecolor{dukeblue}{rgb}{0.0, 0.0, 0.61}
\definecolor{uablue}{rgb}{0.0, 0.2, 0.67}

\renewcommand{\theenumi}{\alph{enumi}}

\definecolor{wheat}{rgb}{0.96,0.87,0.70}

\definecolor{teagreen}{rgb}{0.82, 0.94, 0.75}

\newacronym{hyq}{HyQ}{Hydraulically actuated Quadruped}

\newacronym{lf}{LF}{Left-Front}
\newacronym{rf}{RF}{Right-Front}
\newacronym{lh}{LH}{Left-Hind}
\newacronym{rh}{RH}{Right-Hind}

\newacronym{haa}{HAA}{Hip Adduction-Abduction}
\newacronym{hfe}{HFE}{Hip Flexion-Extension}
\newacronym{kfe}{KFE}{Knee Flexion-Extension}

\newacronym{efr}{EFR}{Extended Feasible Region}
\newacronym{imu}{IMU}{Inertial Measurement Unit}
\newacronym{dofs}{DoFs}{Degrees of Freedom}
\newacronym{rt}{RT}{Real Time}

\newacronym{com}{CoM}{Center of Mass}
\newacronym{cop}{CoP}{Center of Pressure}
\newacronym{zmp}{ZMP}{Zero Moment Point}
\newacronym{icp}{ICP}{Instantaneous Capture Point}
\newacronym{cmp}{CMP}{Centroidal Moment Pivot}
\newacronym{grfs}{GRFs}{Ground Reaction Forces}

\newacronym{ls}{LS}{Least Square}
\newacronym{to}{TO}{Trajectory Optimization}
\newacronym{lipm}{LIPM}{Linear Inverted Pendulum Model}
\newacronym{slip}{SLIP}{Spring Loaded Inverted Pendulum}
\newacronym{eom}{EoM}{Equation of Motions}
\newacronym{qp}{QP}{Quadratic Programming}
\newacronym{sqp}{SQP}{Sequential Quadratic Programming}
\newacronym{slq}{SLQ}{Sequential Linear Quadratic}
\newacronym{mic}{MIC}{Mixed-Integer Convex}
\newacronym{cmaes}{CMA-ES}{Covariance Matrix Adaptation Evolution Strategy}
\newacronym{ara}{ARA*}{Anytime Repairing A*}
\newacronym{pca}{PCA}{Principal Component Analysis}
\newacronym{cpg}{CPG}{Central Pattern Generator}
\newacronym{wbc}{WBC}{Whole-Body Control}

\newacronym{lqr}{LQR}{Linear Quadratic Regulator}
\newacronym{mpc}{MPC}{Model Predictive Control}
\newacronym{nmpc}{NMPC}{Nonlinear Model Predictive Control}
\newacronym{rti}{RTI}{real-time iteration}
\newacronym{ocp}{OCP}{Optimal Control Problem}
\newacronym{irk}{IRK}{Implicit Runge-Kutta}
\newacronym{erk}{ERK}{Explicit Runge-Kutta}
\newacronym{ie}{IE}{implicit Euler}
\newacronym{ee}{EE}{explicit Euler}
\newacronym{ip}{IP}{Iterative Projection}

\newacronym{cwc}{CWC}{Contact Wrench Cone}
\newacronym{awp}{AWP}{Actuation Wrench Polytope}
\newacronym{fwp}{FWP}{Feasible Wrench Polytope}
\newacronym{gws}{GWS}{Grasp Wrench Space}
\newacronym{wfw}{WFW}{Wrench-Feasible Workspace}
\newacronym{fsw}{FSW}{Feasible Solution of Wrench}
\newacronym{FWP}{FWP}{Feasible Wrench Polytope}

\newacronym{cdm}{CD}{Centroidal Dynamics}
\newacronym{nlp}{NLP}{Nonlinear Programming}
\newacronym{ddp}{DDP}{Differential Dynamic Programming}
\newacronym{srbd}{SRBD}{Single Rigid Body Dynamics}
\newacronym{srbm}{SRBM}{Single Rigid Body Model}

\newacronym{stance}{STANCE}{\textbf{S}oft \textbf{T}errain \textbf{A}daptation a\textbf{n}d \textbf{C}ompliance \textbf{E}stimation}

\newacronym{wbopt}{WBOpt}{Whole Body Optimization}

\newacronym{hc}{HC}{Hunt and Crossley's}
\newacronym{kv}{KV}{Kelvin-Voigt's}

\newacronym{wllsr}{WLLSR}{Weighted Linear Least Squared Regression}

\newacronym{mae}{MAE}{Mean Absolute Tracking Error}

\newacronym{ode}{ODE}{Open Dynamics Engine}

\newacronym{vfa}{VFA}{Vision-based Foothold Adaptation}
\newacronym{cnn}{CNN}{Convolutional Neural Network}

\newcommand{\Rnum}{\mathbb{R}} 

\newcommand{\vect}[1]{\mathbf{#1}} 

\newcommand{\mx}[1]{\mathbf{\bm{#1}}} 				

\DeclareMathOperator*{\argmin}{\arg\!\min}				
\newcommand{\mat}[1]{\ensuremath{\begin{bmatrix}#1\end{bmatrix}}}	

\newcommand{\Spd}[1]{\ensuremath{\mathbb{S}_+^{#1}}}			

\hypersetup{draft}
\makeglossaries

\title{Model Predictive Control with Environment Adaptation for Legged Locomotion}
\author{Niraj Rathod$^{1,2}$,  Angelo Bratta$^{2,3}$, Michele Focchi$^2$,  Mario Zanon$^1$, Octavio Villarreal$^2$, Claudio Semini$^2$ and Alberto Bemporad$^1$
	\thanks{$^1$ The authors are with the IMT School for Advanced Studies Lucca, Lucca, Italy {\tt\small \href{mailto:niraj.rathod@imtlucca.it}{name.surname@imtlucca.it}}}
	\thanks{$^2$ The authors are with the Dynamic Legged Systems Lab, Istituto Italiano di Tecnologia (IIT), Genova, Italy.
		{\tt\small \href{mailto:name.surname@iit.it}{name.surname@iit.it}}}
	\thanks{$^3$ Dipartimento di Informatica, Bioingegneria, Robotica 
			e Ingegneria dei Sistemi (DIBRIS), Università di Genova, Genova, Italy}
}

\begin{document}
\begin{flushleft}
\begin{minipage}{\textwidth}\thispagestyle{empty} 
  \includegraphics[scale=.85]{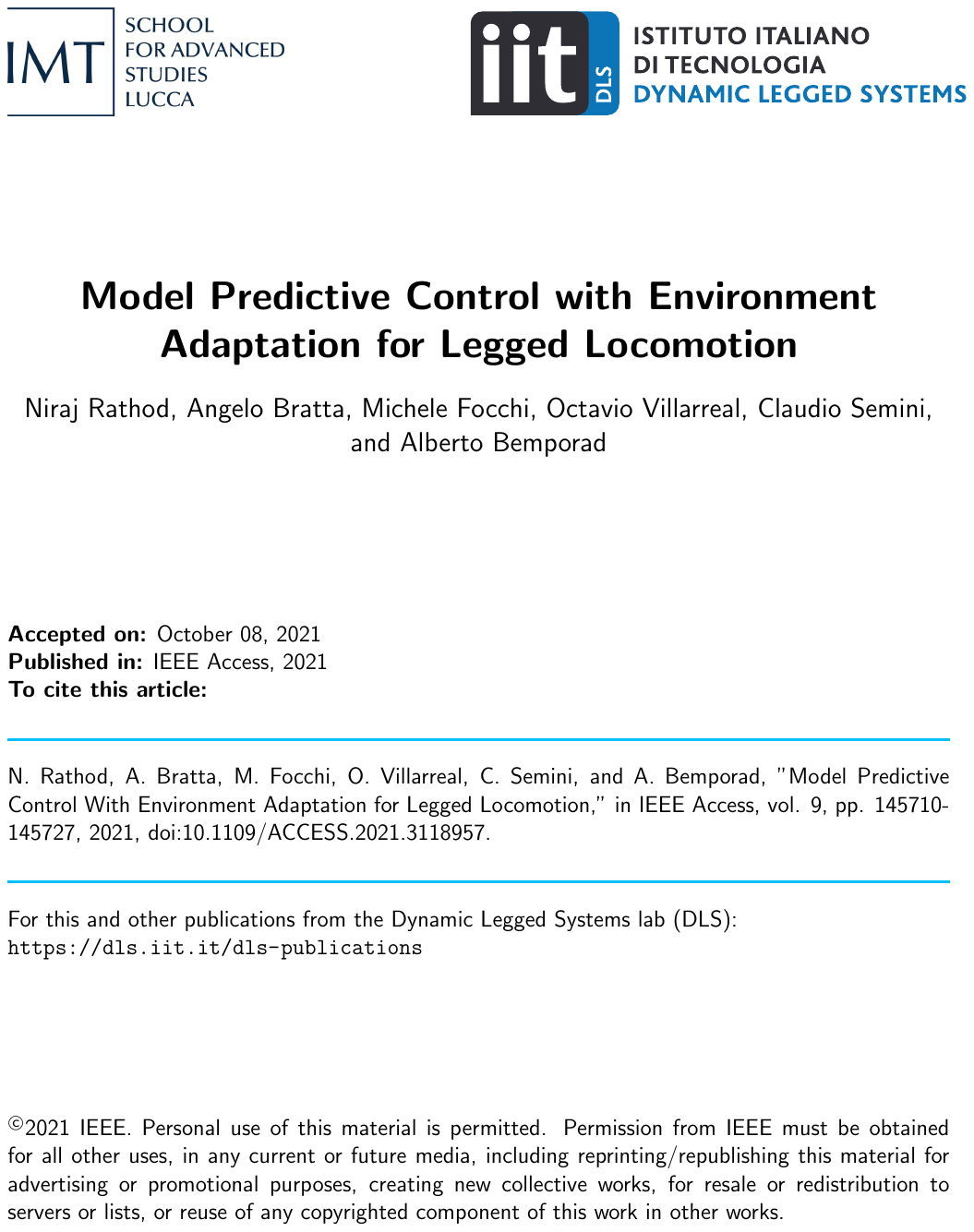}
\end{minipage}
\end{flushleft}

\maketitle
\setcounter{page}{1}
\begin{abstract}
Re-planning in legged locomotion is crucial to track the desired user velocity while adapting 
to the terrain and rejecting external disturbances.
In this work, we propose and test in experiments a real-time  \gls{nmpc} tailored 
to a legged robot for achieving dynamic locomotion on a variety of terrains. 
We introduce a  mobility-based criterion to define an \gls{nmpc} cost that 
enhances the locomotion of quadruped robots while maximizing leg mobility and 
improves adaptation to the terrain features. Our \gls{nmpc} is based on the real-time iteration
scheme that allows us to re-plan online at  $25\,\mathrm{Hz}$ with a prediction horizon of 
$2$ seconds. We use the single rigid body dynamic model defined in the center of mass frame in order 
to increase the computational efficiency. 
In simulations, the \gls{nmpc} is tested 
to traverse a set of pallets of different sizes, to walk into a V-shaped chimney,
and to locomote over rough terrain.
In real experiments, we demonstrate 
the effectiveness of our \gls{nmpc} with the mobility feature that 
allowed IIT's $87\, \mathrm{kg}$ quadruped robot HyQ to achieve an omni-directional walk on flat terrain, 
to traverse a static pallet, and to adapt to a repositioned pallet during a walk. 
\end{abstract}


\section*{Nomenclature}
The list of most commonly used symbols used in this article.
\emph{Acronyms:}

\begin{tabular}{@{} l l @{}}
	CoM & Center of Mass.\\
	GRFs & Ground Reaction Forces.    \\
	NMPC & Nonlinear Model Predictive Control.  \\
	RTI & Real-time Iteration. \\
	SRBD & Single Rigid Body Dynamics.    \\
	VFA & Vision-based Foothold Adaptation. \\
	WBC & Whole-Body Control.   \\
	ZMP & Zero Moment Point.  
\end{tabular}
\label{tab:nomenclature}

\noindent\emph{Notation:}

\begin{tabular}{@{} l l @{}}
	$n_x$ & Number of states.    \\
	$n_u$ & Number of control inputs.    \\
	$n_a$ & Number of model parameters.   \\
	$T$ & Prediction horizon.  \\
	$N$ & Number of control intervals.    \\
	$\mu$ & Friction coefficient. \\
	$\vect{x}^\mathrm{p} \in \Rnum^{n_x \times (N+1)}$ & Predicted states by NMPC. \\
    $\vect{u}^\mathrm{p} \in \Rnum^{n_u \times N}$ & Optimal control inputs from NMPC.
\end{tabular}

\begin{tabular}{@{} l l @{}}
	$\vect{x}^\mathrm{ref} \in \Rnum^{n_x \times (N+1)}$ & Reference states. \\
	$\vect{u}^\mathrm{ref} \in \Rnum^{n_u \times N}$ & Reference control inputs.\\
	$\vect{p}_\mathrm{c}\in \Rnum^3$ & Robot's CoM position.\\
	$\vect{v}_\mathrm{c}\in \Rnum^3$ & Robot's CoM velocity.\\
	$\bm{\Phi}\in \Rnum^3$ & Orientation of robot's base.\\
	$\bm{\omega}\in \Rnum^3$ & Angular velocity of robot's base. \\
	$\vect{f}_i \in \Rnum^3$ & GRF at $i^{th}$ foot. \\
	$\vect{p}_{\mathrm{f},i} \in \Rnum^3$ & Foot position of $i^{th}$ foot. \\
	$\bm{\delta} \in \Rnum^4$ & Contact status. \\
	${}_\mathcal{C}\vect{p}_\mathrm{hf} \in \Rnum^{12}$ & Hip-to-foot distance in CoM frame. \\ 
\end{tabular}
\label{tab:acronyms}
\begin{figure}[t]
	\centering
	\includegraphics[scale=0.14]{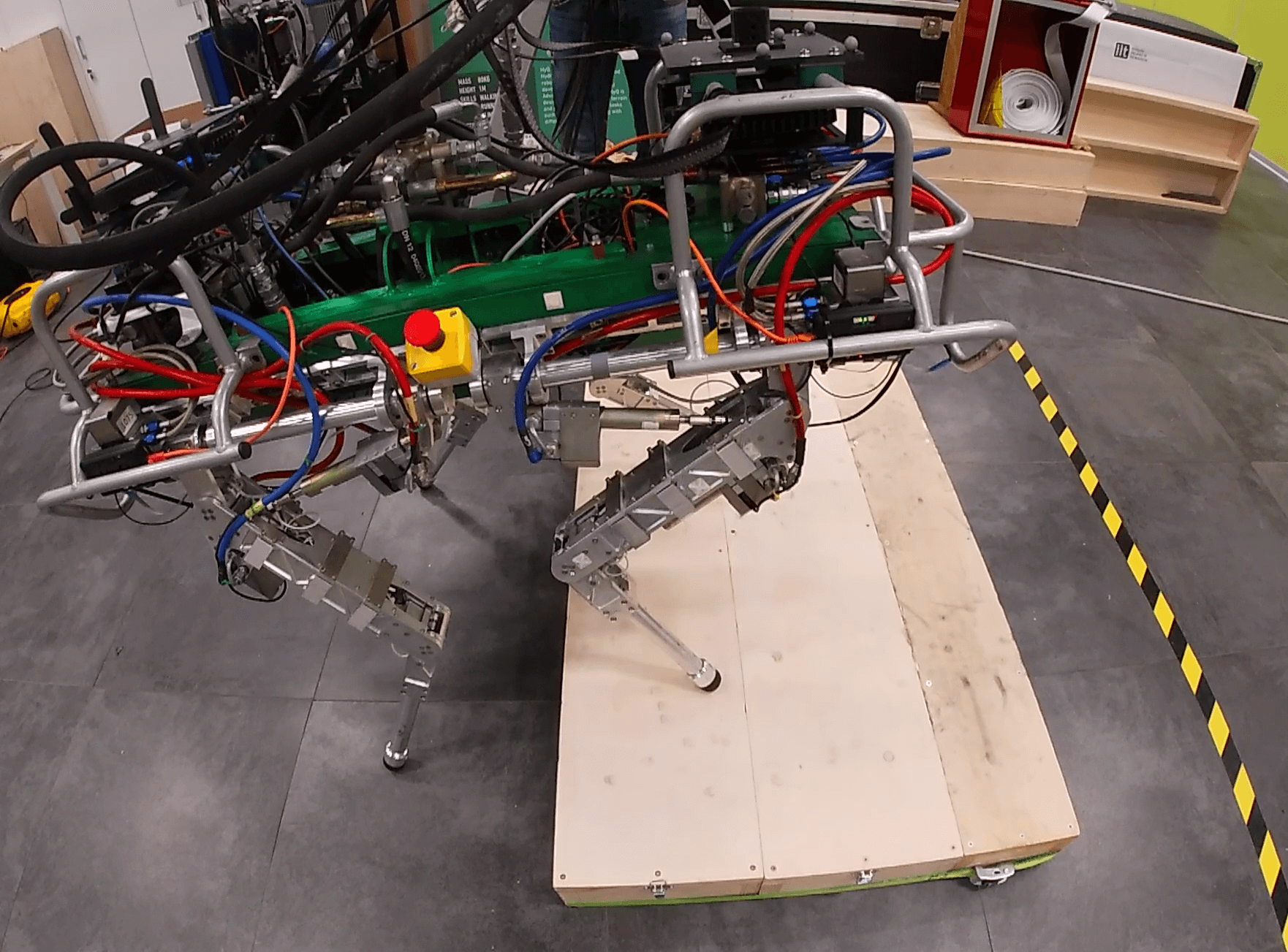}
	\caption{ IIT's quadruped robot HyQ traversing a pallet with the mobility enhanced real-time \gls{nmpc}.}
	\label{fig:pallet_walk_main}
\end{figure}

\section{Introduction}
\label{sec:introduction}
The main advantage of legged robots with respect to their wheeled counterpart is their 
ability to traverse complex and unstructured environment such as forests, obstacles, 
and debris. However, the control of legged robots poses complex problems related to
underactuation (the body is controlled only indirectly through the legs), and to the hybrid 
nature of the forces required to generate motion, since the robot needs 
to establish and interrupt contact between its feet and the ground.
The control design for legged robots was initially dealt with by using heuristic 
approaches which yielded successful results such as in the walking machines from 
Raibert \cite{raibert1986legged}, the virtual model control
of Pratt et al. \cite{pratt2001} and in the heuristic locomotion planning for quadrupedal 
robots by Focchi et al. \cite{Focchi2020}. However,  heuristic approaches have several limitations, for example:
1) they cannot be easily generalized to all kinds of terrain and motions;  2) they 
cannot account for the future state of the robot hence, they have no possibility to guarantee physical 
feasibility of the planned trajectories. The challenge of avoiding 
these undesirable  \textit{myopic} behaviors in heuristic planning 
approaches has motivated the research towards new optimization-based predictive locomotion planning. 

Formulating the locomotion planning as an optimization problem 
allows one to represent high-level locomotion tasks as cost functions and system dynamics 
using constraints. 
Besides robot dynamics, the locomotion tasks should also respect the contact dynamics 
such as unilateral force and friction cone constraints, that 
are critical to stabilize the locomotion.
The use of optimization techniques to design \gls{wbc} has 
enabled legged robots to traverse soft terrains~\cite{Fahmi2020TRO} and to be 
versatile in terms of type of gait and motions that a legged robot
can achieve~\cite{Bellicoso2017}. The aforementioned 
examples are based on the solution of a Quadratic Program 
that only considers the instantaneous \cite{kuindersma2014} effects of the joint 
torques on the robot's base. 
Further, similar to heuristic approaches mentioned earlier, these  approaches
do not consider the information about the future states of the robot and
hence cannot assure recursive feasiblity. 
%
%

In order to address this issue, several approaches make use of \gls{to}-based locomotion 
planning considering the full 
dynamics of the robot~\cite{Neunert2017,Posa2016}.
However, these approaches usually suffer from high computational 
time hence they are often restricted to \textit{offline} (open-loop) use.
In general, offline planners~\cite{Bernardo2018,Melon2020} 
neither adapt to quick terrain 
changes nor cope with state drifts and uncertainties.
%
%
To address this issue the concept of \textit{online} re-planning can be used. 
Online re-planning can intrinsically cope with the problem of error 
accumulation in planned motion that is common in real scenarios.
%
%

For online re-planning, MPC has gained broad interest in 
the robotics community for legged locomotion. 
%
%
Moreover, the intrinsic feedback mechanism offered by MPC can compensate for modeling 
errors and disturbances acting on the system provided that 
the MPC is executed at a sufficiently high rate in closed-loop. 
%
%
A careful choice of the dynamic model inside MPC formulation is 
typically required to achieve a desired re-planning frequency in 
closed-loop, given the limited computational resources available for online computations. 
For example, using a full dynamics model 
of legged robots inside MPC \cite{Neunert2018, koenemann15iros} with long prediction horizon may result in an optimization
problem which requires excessive computations for real-time deployment at high sampling rates.
Using approximate models is a way to reduce the complexity of the 
optimization, trading the accuracy with computational efficiency.
%
%
Following this line, \cite{Dai2014} used a \gls{cdm} plus a full-kinematic 
model to enforce the kinematic limits in \gls{to} to plan complex behaviors on the humanoid robot Atlas. 
%
%
The \gls{cdm} model considers contact forces as input and
links the linear and angular momentum of the 
robot to the external wrench~\cite{Orin2013a}.

A simplified version of the \gls{cdm} model is the \gls{srbd} model where the inertia of the legs is neglected (assumption of massless legs) 
and the robot's body and legs are lumped into a single rigid body. 
This model is well suited for quadrupeds, since they 
usually concentrate their mass and inertia in the robot base, unlike humanoids.
%
%
The \gls{srbd} model was used for \gls{to}~\cite{Winkler2018a}
and  MPC~\cite{wensigHeuristics2017} to jointly optimize 
for footholds, \gls{com} trajectories and contact forces.
%
%
By further linearizing the angular part of the 
dynamics  of the \gls{srbd}, \cite{DiCarlo2018} was able to 
achieve a variety of quadrupedal gaits in experiments 
but their approach was not suitable for motions that involve
large variations from the horizontal orientation. 
%
%
The simplest among all the approximate models mentioned earlier is the \gls{lipm} and it has been used inside MPC for quadruped~\cite{Horvat2017} and biped~\cite{herdt2015} locomotion. 
However, there are two main limitations in \gls{lipm}, 
namely it neglects  angular dynamics and assumes constant robot height. 
Additionally, it does not account for friction cones, 
so that the contact stability 
on non-flat terrain  can not be guaranteed. 


While models play an important role in obtaining computationally light MPC formulations,
the choice of solution method is also paramount to achieve fast online re-planning with MPC.
A \gls{ddp} based approach demonstrated the real-time performance
with  whole-body MPC \cite{koenemann15iros} on HRP-2 humanoid. 
Recently, \cite{Grandia2019} proposed a \gls{ddp}-based MPC using a kinodynamic model 
which re-plans at a frequency of  $15\,\mathrm{Hz}$ with a prediction horizon of  $1\,\mathrm{s}$ on a quadruped.
The main drawbacks of \gls{ddp} based approaches is the difficulty in implementing hard-inequality and switching  constraints. 
Hard-inequality constraints need to be  implemented as penalties (e.g., with  relaxed-barriers) 
\cite{Hauser2007} while the switching constraints are formulated using Augmented Lagrangian methods 
\cite{Lantoine2012, Li2020}. Though not obvious at first sight, these methods are essentially equivalent to direct optimal control based on multiple-shooting~\cite{Bock1984} in combination with some form of \gls{nlp} solvers using barrier functions in the real-time iteration scheme~\cite{Diehl2005}. One such framework is provided by $\texttt{acados}$~\cite{Verschueren2019}.

In addition to \gls{ddp}-based MPC, there also exist a few implementations of \gls{nlp}-based MPC for legged robots. 
One such \gls{nmpc} implementation with \gls{com} dynamics plus full kinematic model was demonstrated 
in \cite{Farshidian2017b} using a \gls{slq} algorithm for a trotting gait 
on flat terrain. Neunert  et al. \cite{Neunert2018} achieved a fast re-planning 
frequency of  $80$-$170~\mathrm{Hz}$ for a small prediction horizon of  $0.5~\mathrm{s}$ ($125$ nodes)
with their \gls{nmpc} using the full dynamics of the robot, and
optimizing foot locations, swing timing, and locomotion sequences along 
with full body dynamics.  However, in the real experiments they have only demonstrated slow trotting on flat terrain. 
Moreover, since their approach does not consider the map of the terrain, it has  limited 
application on uneven terrain conditions.  An interesting observation is that 
they did not see a noticeable degradation in the closed-loop performance of the \gls{nmpc} 
when the re-planning frequency was dropped until 30 $\mathrm{Hz}$, demonstrating
that the predictive nature of the  MPC empowers the robot to tolerate much lower 
re-planning frequency. A similar observation was made in \cite{Bledt2019} with 
an MPC scheme which optimizes foot locations, but requires a heuristic conditioning 
of the cost function. In their experiments, the robot is stable if the re-planning occurs at $20 \ \mathrm{Hz}$, 
unstable for lower frequencies and the performance improvement is observed over 
$40 \ \mathrm{Hz}$.

%
%


The aforementioned approaches have been successful in controlling legged robots, 
but neglected an important aspect of these robots, which is usually referred to 
as \emph{mobility}. In this paper, we define the mobility as the attitude
of the robot leg to arbitrarily change its foot position~\cite{sciavicco00}. We noticed that maximizing mobility improves 
terrain adaptation hence it is advantageous to account for it in the motion planning of legged robots.
Furthermore, as discussed in Section \ref{subsec:mobility}, adding mobility in the \gls{nmpc} cost
eliminates the need to specify references for the roll, pitch and height of the robot.

To achieve kinematically suitable configurations for the legs, a common heuristic 
is to align the robot base with the terrain 
inclination (estimated in \cite{Focchi2020} via 
fitting an averaging plane through the stance feet). 
This approach aims to bring the legs as close as possible to the 
middle of their workspaces in order to avoid the violation of the kinematic limits. 
Optimization of mobility allows to achieve a similar behaviour in an automatic way.
Fankhauser et al. in \cite{Fankhauser2018}
maximized mobility by encoding it in a cost function that
penalizes the distance with respect to a default foot position. Recently, 
Cebe et al. \cite{Cebe2020} implemented \gls{to} using an \gls{srbd}
model and also incorporating the feet positions in the optimization.
They re-plan only at the feet touchdowns due to the high computation demand of 
their \gls{to} algorithm and showed experimental results on uneven
terrain. Since their planner does not plan during the swing phase of the legs, 
they do not run their planner in an MPC fashion. Apart from the previously mentioned contributions, 
to the best of our knowledge no prior work has addressed the mobility 
with MPC in legged locomotion.

\subsection{Proposed Approach and Contribution}
In this work, we demonstrate in experiments with 
our $87 \, \mathrm{kg}$ \gls{hyq} robot \cite{semini11hyqdesignjsce} that a suitably formulated 
NMPC can tackle rough terrain locomotion, account for leg mobility, and
provide the optimal base orientation, while being real-time feasible.
Indeed, optimizing for leg mobility allows our \gls{nmpc} to devise a robot base orientation and height that improves locomotion on rough terrain.  

This is particularly useful to achieve \textit{environment adaptation} on rough terrains.
Another advantage is that   minimal heuristics is  required from the 
user i.e., no reference trajectory for the robot's height, and its base roll and pitch orientation
is needed.

This work is a system integration on the same line of our previous work \cite{Mastalli20TRO}. 
However, while in \cite{Mastalli20TRO} only offline optimization was performed, 
here we achieve real-time feasible online replanning in an MPC fashion.
To achieve this goal: 
\begin{enumerate}
	\item  We consider a simplified \gls{srbd} model
that describes the angular and translational dynamics 
of the robot base but neglects the dynamics of legs. 
	\item We employ the \gls{rti} scheme \cite{Diehl2005,Diehl2009,Gros2020} 
that allows us to run our \gls{nmpc} online with the prediction horizon of $2\,\mathrm{s}$ (50 nodes)
as opposed to the $0.5 \,\mathrm{s}$ ($125$ nodes) used by \cite{Neunert2018}. Differently from
\cite{Cebe2020} (that re-plans at each foot touchdown event),
we continuously re-plan at the rate of $25\,\mathrm{Hz}$.
\item We run our \gls{nmpc} on a single computer along with the rest of our locomotion 
framework\footnote{Except the perception related modules that run on a dedicated computer}
unlike in \cite{Cebe2020, Neunert2018} where they use dedicated computers 
for their \gls{to} and \gls{nmpc}, respectively.
\end{enumerate}

We show in simulation the robot traversing a set of 
pallets of different dimensions placed relatively at varying distances, 
walking into a V-shaped chimney and lastly over a randomly generated rough terrain. 
We present \emph{Experimental} results that demonstrate the capability of our \gls{nmpc} to generate an 
omni-directional walk and to traverse a pallet for our quadruped 
robot \gls{hyq} (see Fig. \ref{fig:pallet_walk_main}). We tested the re-planning capability of our approach 
by pushing a pallet in front of the robot while walking, such that the control 
algorithm has to re-plan online in order to adapt to a dynamically changing environment.
\begin{figure*}[htb]
	\centering
	\includegraphics[width=\textwidth]{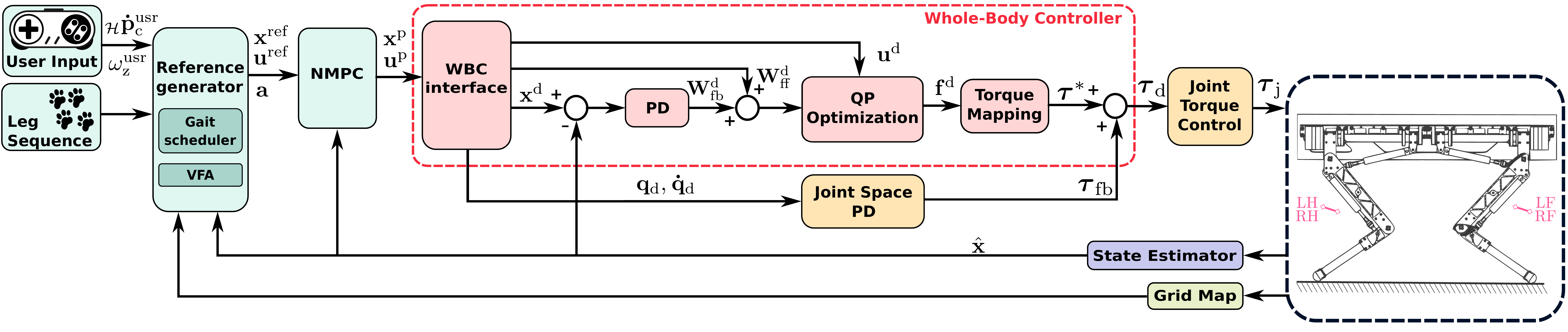}
	\caption{ Block diagram of the planning pipeline with the \gls{nmpc} 
		in our locomotion framework. The reference generator provides the 
		references ($\vect{x}^{\mathrm{ref}}$, $\vect{u}^{\mathrm{ref}}$) 
		to NMPC after receiving the user inputs. Then the NMPC passes 
		optimal state $\vect{x}^{\mathrm{p}}$ and control $\vect{x}^{\mathrm{p}}$ 
		trajectories to the Whole-Body Controller. The torque $\bm{\tau}_\mathrm{d}$ 
		is given as reference to low level joint torque controller. The state estimator
		provides the state estimation $\vect{\hat{x}}$ to the required blocks.
		Finally, the heightmap is generated by Grid Map and given to the reference generator.}
	\label{fig:blockDiagram}
\end{figure*} 
To summarize, the contributions of our work (in order of importance) are as follows:
\begin{itemize}[leftmargin=*]
	
	\item Major \textit{experimental}  results are presented in Section \ref{sec:results} 
	where we demonstrate the capability of our NMPC planner
	to generate an omni-directional walk and rough
	terrain locomotion for our quadruped robot HyQ, by exploiting
	an online evaluation of the map of the terrain and using on-board state estimation. 
		
	\item Additional (minor) contributions are: 
	\begin{enumerate}		
	\item We introduce a cost term which accounts for 
	\textit{mobility} by penalizing hip-to-foot positions 
	that do not provide the highest mobility. 
	To the best of our knowledge, this is the first time that mobility has been addressed with MPC.
	\item The generation of the reference trajectory for the \gls{nmpc} takes 
	into account premature and delayed touchdown of the feet as 
	well as continuously adjusting the footholds according to the robot body motion 
	and the terrain features. 
	\item We use a parametric robot model that results in smaller NMPC formulation.
	\end{enumerate}
\end{itemize}

\subsection{Outline}
The paper is organized as follows: Section \ref{sec:overview}
gives an overview of our planning pipeline whereas 
Section \ref{sec:nmpc} describes the  \gls{nmpc} setup. 
The leg mobility and other features 
are explained in Section \ref{sec:features},
whereas the generation of the references and the \gls{wbc} are detailed 
in Sections \ref{sec:ref_gen} and \ref{sec:whole_body_controller}, respectively. 
We then summarize the \gls{rti} scheme for our \gls{nmpc} in Section \ref{sec:rti_scheme}. 
Further, Section \ref{sec:results} illustrates simulation and experimental results 
with the \gls{hyq} robot. Finally, we draw the conclusions in Section \ref{sec:conclusion}.

\section{Locomotion Framework} \label{sec:overview}
%
%
Fig.~\ref{fig:blockDiagram} illustrates the planning pipeline of our 
locomotion framework. The \emph{reference generator}, as discussed in Section 
\ref{sec:ref_gen}, takes the user input (longitudinal, lateral and angular velocity), 
schedule of the gait (e.g., a crawl) timing, the initial state of the robot, and a
map of the terrain to generate reference trajectories for the state $\vect{x}^{\mathrm{ref}}$ 
and control input $\vect{u}^{\mathrm{ref}}$ required by the \gls{nmpc}.
The reference generator also provides a vector of parameters $\vect{a}$ to the \gls{nmpc},
that includes foot locations and sequences of contact status.
The \gls{nmpc} running at  $25 \,\mathrm{Hz}$,
delivers the optimal trajectories of the  
state $\vect{x}^\mathrm{p}$ and control input 
$\vect{u}^\mathrm{p}$, as detailed in Section \ref{sec:nmpc}. 

All the components 
of the Whole-Body controller (highlighted with dashed box in Fig.~\ref{fig:blockDiagram}) are discussed in Section \ref{sec:whole_body_controller}. 
The \emph{\gls{wbc} interface}  interpolates the optimal state $\vect{x}^\mathrm{p}$ at a 
rate of  $250 \,\mathrm{Hz}$ to generate a desired  signal $\vect{x}^\mathrm{d}$ 
for a Cartesian virtual impedance controller \cite{Focchi2016}. 
The \gls{wbc} interface also  computes the feedforward wrench 
$\vect{W}^\mathrm{d}_\mathrm{ff}$ that is added to a feedback wrench 
$\vect{W}^\mathrm{d}_\mathrm{fb}$ that renders   
the Cartesian  impedance.
Moreover, the \gls{wbc} interface provides  the joint position $\vect{q}_\mathrm{d}$
and velocities $\vect{\dot{q}}_\mathrm{d}$ to a Joint Space PD controller running at  $1\,\mathrm{kHz}$. 
After acquiring the feedback and 
feed-forward wrenches, a \gls{qp} optimization computes the  
vector of desired \gls{grfs} $\vect{f}^\mathrm{d}$ accounting for the friction cone constraints 
and penalizing the the difference between $\vect{f}^\mathrm{d}$ and 
$\vect{u}^\mathrm{p}$ coming from the \gls{nmpc} solution. 
Then, $\vect{f}^\mathrm{d}$
is mapped to the torque vector $\bm{\tau}^*$ that is  added to 
the Joint Space PD torques $\bm{\tau}_\mathrm{fb}$ resulting into the total  
desired torque $\bm{\tau}_\mathrm{d}$. Ultimately, $\bm{\tau}_\mathrm{d}$ is  passed to a low-level joint torque controller 
as reference \cite{boaventura15tro}. 

An online  state estimator~\cite{nobili_camurri2017rss} that runs at  $500\,\mathrm{Hz}$ provides the estimation of the 
robot state $\vect{\hat{x}}$ to all the  components inside 
our locomotion framework that require it. 
A dedicated on-board computer takes inputs from an RGB-D camera (RealSense) 
mounted in front of the robot and generates a 2.5D heightmap  at the rate
of  $30\,\mathrm{Hz}$ using the \emph{Grid Map} library from \cite{Fankhauser2016GridMapLibrary}. 
This heightmap is later sent to the reference generator. 
\section{NMPC}\label{sec:nmpc}
In our planning algorithm, we choose a real-time \gls{nmpc} formulation because it has the ability to handle 
both the nonlinear system dynamics and the constraints, explicitly. 
\gls{nmpc} is based on solving an \gls{ocp} given the current 
state $\vect{\hat{x}}_0$ of the system. 
Only the first element of the optimized input trajectory 
is applied to the system, then the state is measured and the \gls{ocp} is solved again
based on the new state measurement to close the loop. 

We define the decision variables as the predicted state and control input with $\vect{x}^\mathrm{p}=\{\vect{x}_{0}, \ldots, \vect{x}_N\}$ and  $\vect{u}^\mathrm{p}=\{\vect{u}_{0}, \ldots, \vect{u}_{N-1}\}$, respectively, 
such that an \gls{nlp} formulation can be stated as:
\begin{subequations}\label{eq:nlp_formulation}
	\begin{align}
	\displaystyle{\min_{\vect{x}^\mathrm{p},\vect{u}^\mathrm{p}}}\quad
	&  \sum_{k=0}^{N-1} \ell\left(\vect{x}_{k}, \vect{u}_{k}, \vect{a}_{k}\right)+\ell_\mathrm{T}\left(\vect{x}_{N}\right) \label{eq:cost_function} \\
	\text{s.t.} \quad 		& \vect{x}_{0}=\boldsymbol{\hat{\vect{x}}_{0}}, \label{eq:initial_condition}\\
	&\vect{x}_{k+1}=f\left(\vect{x}_{k}, \vect{u}_{k}, \vect{a}_{k}\right),&& k\in\mathbb{I}_0^{N-1}, \label{eq:equality}\\
	&h\left(\vect{x}_{k}, \vect{u}_{k}, \vect{a}_{k}\right) \leq 0,&& k\in\mathbb{I}_0^{N-1}, \label{eq:inequality}
	\end{align}
\end{subequations}
where, $\ell : \Rnum^{n_x} \times \Rnum^{n_u} \times \Rnum^{n_a} \rightarrow \Rnum$ is the stage 
cost function; $\ell_\mathrm{T} : \Rnum^{n_x} \rightarrow \Rnum$ is the terminal 
cost function. The initial condition \eqref{eq:initial_condition} 
is expressed by setting $\vect{x}_0$ equal to the state estimate 
$\boldsymbol{\hat{\vect{x}}_{0}}$ received from the state estimator. 
The vector of model parameters $\vect{a}_{k}$ is not optimized 
but it is computed externally by the reference generator and 
provided to the optimization problem formulation. The nonlinear system 
dynamics are introduced by the equality constraints \eqref{eq:equality}. 
Finally, the path constraints are included with \eqref{eq:inequality} 
which, for example, can be bounds on the decision variables.
The \gls{nlp} \eqref{eq:nlp_formulation} is defined
for a \emph{prediction horizon} $T$ that is divided into $N$ discrete time \emph{control intervals} 
of lengths $T_\mathrm{s} = \frac{T}{N}$. Hereafter, we will refer to $T_\mathrm{s}$ 
as the \textit{sampling time}.

\subsection{Cost}
In our \gls{nmpc} formulation we use a cost function of the form:
\begin{subequations}\label{eq:quadratic_cost}
	\begin{align}
	\ell\left(\vect{x}_{k}, \vect{u}_{k},\vect{a}_k\right)=& \,\ell_\mathrm{t} + \ell_\mathrm{m} + \ell_\mathrm{r}, \\
	\ell_\mathrm{t} =&\parallel \vect{x}_k-\vect{x}^{\mathrm{ref}}_k\parallel_{\mx{Q}}^2+
	\parallel \vect{u}_k-\vect{u}^{\mathrm{ref}}_k\parallel_{\mx{R}}^2, \label{eq:tracking_cost}\\
	\ell_\mathrm{m} = &\parallel {}_\mathcal{C}\vect{p}_{\mathrm{hf}_{k}}-{}_\mathcal{C}\vect{p}_{\mathrm{hf}_{k}}^{\mathrm{ref}}\parallel_{\mx{M}}^2, \label{eq:mobility_cost}\\
	\ell_\mathrm{r} = &\,\rho\parallel {{}_\mathcal{K}\vect{u}_k}\parallel_{\mx{P}}^2\label{eq:force_robustness}
	\end{align}
\end{subequations}
%
%
%
\begin{itemize}[leftmargin=*]
	\item The tracking cost \eqref{eq:tracking_cost} is associated to state and control input 
	and the references trajectories $\vect{x}^{\mathrm{ref}}_k, \vect{u}^{\mathrm{ref}}_k$ 
	are provided by the reference generator for each sampling instance $k$ (refer Section \ref{sec:ref_gen}).
	\item The mobility cost \eqref{eq:mobility_cost} is one of the contributions 
	of this work that accounts for improving the leg mobility 
	by penalizing the difference between the hip-to-foot distance 
	${}_\mathcal{C}\vect{p}_\mathrm{hf}$ and the reference value ${}_\mathcal{C}\vect{p}_\mathrm{hf}^{\mathrm{ref}}$ 
	of maximum mobility. This cost allows the  \gls{nmpc} to optimize the robot base 
	orientation (e.g. align it to the terrain shape) in order to  increase 
	the leg mobility which has as a desirable consequence to stay far from kinematic limits during locomotion. 
	The derivation of ${}_\mathcal{C}\vect{p}_\mathrm{hf}^\mathrm{ref}$ is detailed 
	separately in Section \ref{subsec:mobility}. 
	\item In some locomotion scenarios \cite{Focchi2016}, to cope with uncertainties 
	in the contact normal estimation and increase robustness to external disturbances, 
	it is desirable to have the \gls{grfs} $\vect{f}_i$ as close as possible to 
	the center of the friction cone. This can be achieved with
	by penalizing $X$-$Y$ components of $\vect{u}$ in a frame $\mathcal{K}$ (see Fig. \ref{fig:hyqScheme}) that
	is aligned to the normal of the contact and it is included in our cost function 
	by a control input regularization term \eqref{eq:force_robustness}, refer Section \ref{subsec:force_robustness} for the details. 
\end{itemize}
The positive definite weight matrices $\mx{Q} \in \Spd{n_x}, \mx{R} \in \Spd{n_u}$, 
$\mx{M} \in \Spd{12}$, $\mx{P} \in \Spd{n_u}$ act as important tuning parameters 
in the \gls{nmpc} formulation. The regularization factor $\rho$ decides the trade-off 
between force robustness cost \eqref{eq:force_robustness} and both the tracking  
\eqref{eq:tracking_cost} and mobility \eqref{eq:mobility_cost} cost. Finally, we define
the terminal cost $\ell_\mathrm{T} = \parallel \vect{x}_N-\vect{x}^{\mathrm{ref}}_N\parallel_{\mx{Q}_N} $ and use the weight matrix ${\mx{Q}_N} ={\mx{Q}}$ for this cost.
%
%
\subsection{Robot Model} \label{sec:robot_model}

\begin{figure}[h]
	\centering
	\includegraphics[scale=0.3]{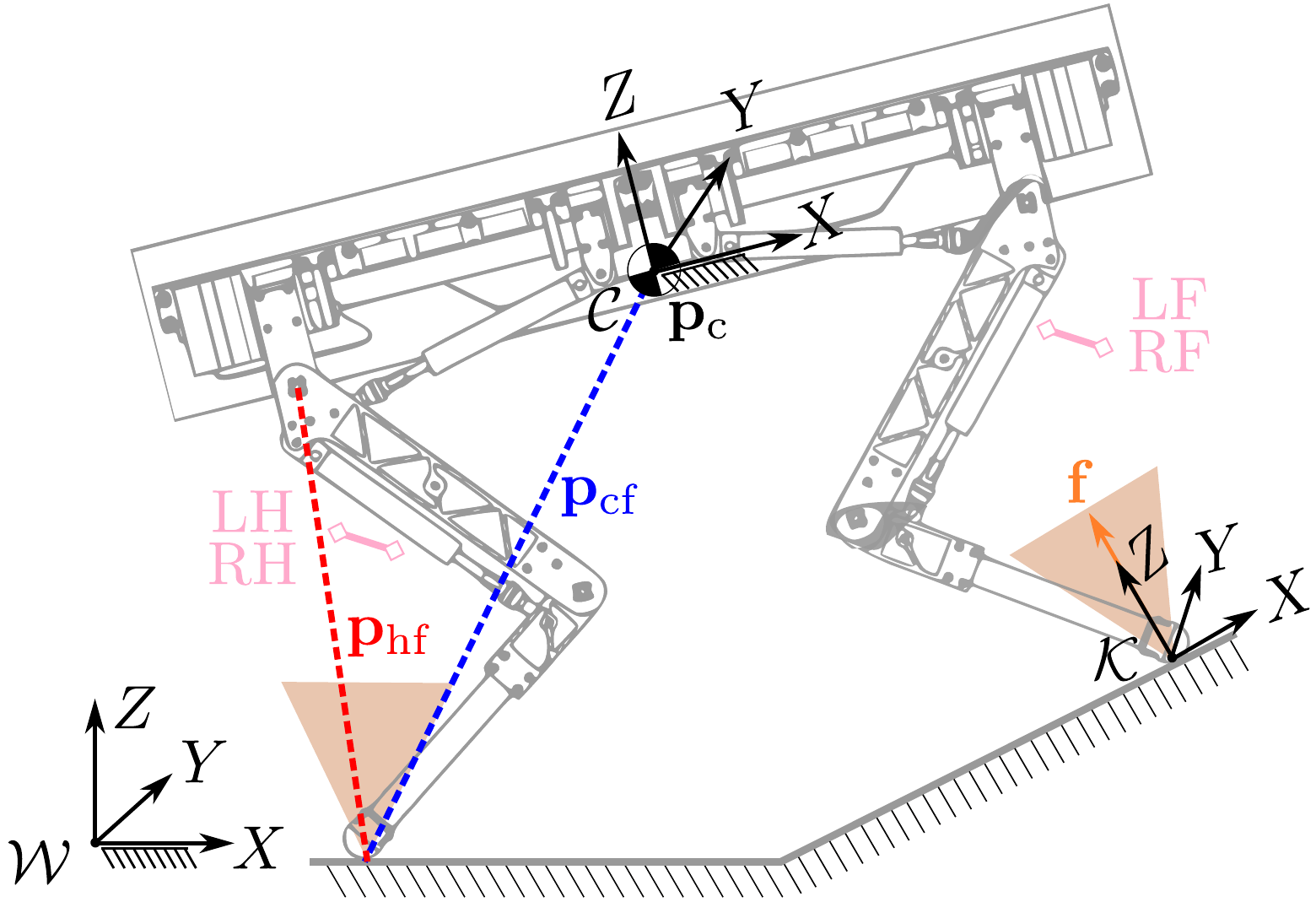}
	\caption{ \gls{hyq} schematic showing the inertial frame ($\mathcal{W}$), the 
		\gls{com} frame ($\mathcal{C}$) attached to the \gls{com} of the robot, and the contact frame ($\mathcal{K}$). The robot legs are shown in the \textit{default} configuration.} 
	\label{fig:hyqScheme}
\end{figure}

The inertial frame $\mathcal{W}$ and the \gls{com} frame $\mathcal{C}$ are shown in Fig.~\ref{fig:hyqScheme}. The \gls{com} frame is aligned with the base of the robot and 
its origin is located at the \gls{com}. A variable with left subscript denotes its 
frame of reference. For example  ${}_\mathcal{C}\bm{\omega}$ represents 
the angular velocity of the robot base
expressed in the \gls{com} frame $\mathcal{C}$. 
Note that, unless explicitly specified,  all the relevant quantities in this paper 
are defined in the inertial frame $\mathcal{W}$. Throughout this paper we define $(a,\ldots,b)$ as the column vector stacking any generic column vectors $a,\ldots,b$.
We use a simplified reduced-order \gls{srbd} model \cite{Winkler2018a} defined in a 6D space that describes 
the translational and angular dynamics of the robot while neglecting 
the dynamics of its swinging legs. This is a valid approximation for the \gls{hyq} robot  because most 
of its mass is concentrated in the base, as mentioned in~\cite{semini11hyqdesignjsce}
(the mass of the base is $61\, \mathrm{kg}$ and the mass of each leg is $6.5\, \mathrm{kg}$).
The robot is approximated as a rigid body with the inertia 
computed considering the robot in a default leg configuration as shown in Fig. \ref{fig:hyqScheme}. 
We choose to define the \gls{srbd} model in the \gls{com} frame (specifically the 
angular dynamics) because this choice yields a constant inertia tensor. 
Thus, the angular dynamic equations are much simpler i.e., 
less non-linear because the inertia tensor is not time varying.
In the \gls{srbd} model, \gls{grfs} are applied as inputs to 
control the position and orientation of the robot base. 
The \gls{srbd} model is:
\begin{subequations} \label{eq:newtonEuler}
	\begin{align}
	m\dot{\vect{v}}_\mathrm{c}&=m\vect{g}+\sum_{i=1}^{4}\delta_i\vect{f}_i \label{eq:trans_dynamics}\\
	{}_\mathcal{C}\mx{I}_\mathrm{c}\,{}_\mathcal{C}\dot{\bm{\omega}}+{}_\mathcal{C}\bm{\omega} \times {}_\mathcal{C}\mx{I}_\mathrm{c} {}_\mathcal{C}\bm{\omega} &=\sum_{i=1}^{4} \delta_i   {}_\mathcal{C}\vect{p}_{\mathrm{cf},i} \times {}_\mathcal{C}\vect{f}_i \label{eq:ang_dynamics}
	\end{align}
\end{subequations}
where $m$ is the robot mass, $\dot{\vect{v}}_\mathrm{c} \in \Rnum^3$ is the \gls{com} acceleration, 
$\vect{g}$ is the gravitational acceleration, 
$\vect{f}_i \in \Rnum^3$ is the ground reaction force at foot $i$, ${}_\mathcal{C}\mx{I}_\mathrm{c}\in \Rnum^{3\times3}$ 
is the inertia tensor computed at the \gls{com} frame origin, ${}_\mathcal{C}\dot{\bm{\omega}}\in \Rnum^3$ 
is the angular acceleration of the robot's base, $\vect{p}_{\mathrm{cf},i}\in \Rnum^3$ is the distance between 
the \gls{com} position $\vect{p}_\mathrm{c} \in \Rnum^3$ and the position $\vect{p}_{\mathrm{f},i}\in \Rnum^3$ of foot $i$. We introduce binary parameters $\delta_i = \{0, 1\}$ to define
whether foot $i$ is in contact with the ground and can therefore generate contact 
forces or not. 

%
%
The robot dynamics governed by \eqref{eq:newtonEuler} can be 
expressed as the continuous-time state-space model:
%
%
\begin{equation}\label{eq:stateSpaceEq}
\mat{\dot{\vect{p}}_\mathrm{c}\\
	\dot{\vect{v}}_\mathrm{c}\\
	\dot{\bm{\Phi}}\\
	{}_\mathcal{C}\dot{\bm{\omega}}}
=
\mat{ {\vect{v}}_\mathrm{c} \\
	1/m \sum_{i=1}^4 \delta_i \vect{f}_i + \vect{g} \\
	\mx{E}'^{-1}(\bm{\Phi}) {}_\mathcal{C}\bm{\omega} \\ 
	-{}_\mathcal{C}\mx{I}^{-1}_\mathrm{c}({}_\mathcal{C}\bm{\omega} \times {}_\mathcal{C}\mx{I}_\mathrm{c}) {}_\mathcal{C}\bm{\omega} +  \sum_{i=1}^4 \delta_i {}_\mathcal{C}\mx{I}^{-1}_\mathrm{c}{}_\mathcal{C}\vect{p}_{\mathrm{cf},i} \times {}_\mathcal{C}\vect{f}_i
}
\end{equation}
where $\vect{v}_\mathrm{c}$ is the \gls{com} velocity of the robot.  The robot base orientation is represented by the sequence of $Z$-$Y$-$X$ Euler 
angles \footnote{Note that Euler angles can suffer from singularities that occur in certain configurations \cite{Younes2012}. Because in this work we do not consider motions that involve such configurations, using Euler angles does not pose any issue. A singularity-free implementation is out of the scope 
of this work and is left for future research.} \cite{Diebel2006} $\bm{\Phi}=(\phi,\, \theta,\, \psi)$ i.e., roll
($\phi$), pitch ($\theta$) and yaw ($\psi$), respectively. The relation between the Euler Angles 
rates $\dot{\bm{\Phi}}$ and angular velocity ${}_\mathcal{C}\bm{\omega}$ is well-known and discussed in Appendix~\ref{adx:eulerRate} for the sake of completeness. We define the state and control vectors as
$\vect{x}=(\vect{p}_\mathrm{c}, \,\vect{v}_\mathrm{c} ,  \, \bm{\Phi}, \, {}_\mathcal{C}\bm{\omega})$, and 
$\vect{u}=(\vect{f}_1, \ldots, \vect{f}_4)$. 
Equation \eqref{eq:stateSpaceEq} 
can be concisely written as:
\begin{equation}\label{eq:compactStateSpace}
\dot{\vect{x}}(t)= g(\vect{x}(t),\vect{u}(t),\vect{a}(t)),
\end{equation}
where $\vect{a} = (\vect{p}_\mathrm{f}, \, \bm{\delta})$
is a  vector of parameters that includes the feet positions $\vect{p}_\mathrm{f}$ and the contact status 
$\bm{\delta} \in \Rnum^4$.

The rigid-body dynamics \eqref{eq:compactStateSpace} are discretized using numerical integration~\cite{Quirynen2017,Butcher2003,Hairer1993,Hairer1996} to obtain the
discrete-time model:
\begin{align}
\vect{x}_{k+1}=f\left(\vect{x}_{k}, \vect{u}_{k}, \vect{a}_{k}\right),
\end{align}
which defines equality constraints~\eqref{eq:equality} imposed at every stage $k$ in MPC to ensure that the state trajectory satisfies the system dynamics for the given control inputs. 



%
%
One specific feature of legged robots is the need to ensure that the values of the \gls{grfs} equal to zero for a swinging leg. This is typically done by introducing complementarity constraints \cite{Cebe2020, Villarreal2020}. These constraints, however, pose several difficulties in the solution of the optimization problem, since the vast majority of the \gls{nlp} algorithms cannot handle them and tailored solvers are required. Ultimately, this results in a significant increase in computation time.
An alternative to complementarity constraints consists in providing the sequence of contact status $\bm{\delta}$ as input parameters in the
state space model \eqref{eq:compactStateSpace}.  
In this manner, a contact mode $\delta_i$ 
is multiplied with the terms involving force $\vect{f}_i$ in \eqref{eq:stateSpaceEq} 
and the contribution of that force is nullified during the swing 
phase of the corresponding leg $i$. 
Hence, there is no more need to include complementarity constraints 
separately in \eqref{eq:nlp_formulation} which results in 
fewer constraints and, consequently, in a relatively smaller \gls{nmpc} formulation. 
%
%

%
%

\subsection{Friction cone and unilateral constraints}

Friction cone constraints are encoded with their square pyramid approximation: 
\begin{subequations}\label{eq:friction_cone}
	\begin{align} 
	-\mu_i \vect{f}_{\mathrm{z},i}&\leq \vect{f}_{\mathrm{x},i}\leq \mu_i \vect{f}_{\mathrm{z},i}\\
	-\mu_i \vect{f}_{\mathrm{z},i}&\leq  \vect{f}_{\mathrm{y},i}\leq \mu_i \vect{f}_{\mathrm{z},i}\\
	\underline{\vect{f}}_\mathrm{z}&\leq \vect{f}_{\mathrm{z},i}\leq \overline{\vect{f}}_\mathrm{z} \label{eq:unilaterality}
	\end{align}
\end{subequations}
where, $\underline{\vect{f}}_\mathrm{z}$ and $\overline{\vect{f}}_\mathrm{z}$ 
are upper and lower bounds on \gls{grfs} $Z$ component, respectively, and
$\mu_i$ is the friction coefficient of the contact surface. 
Choosing 
$\underline{\vect{f}}_z$ greater than or equal to zero enforces unilateral
constraints on the normal forces $\vect{f}_\mathrm{z}$. The friction cone and unilateral constraint are represented by $h\left(\vect{x}_{k}, \vect{u}_{k}, \vect{a}_{k}\right) \leq 0$ in the \gls{nmpc} formulation.
\section{Locomotion-Enhancing Features} \label{sec:features}

In this section we discuss the main distinctive features of our approach, which we found relevant to improve locomotion ability of our quadruped robot. 
These features are \emph{mobility}, \emph{force robustness} and \emph{\gls{zmp} margin}.

%
\subsection{Mobility and Mobility Factor} \label{subsec:mobility}
Terrain adaptability is vital when it comes to locomotion of the legged robots. 
Adjusting the posture of the robot depending on the environment is important for safe locomotion.
A way to enable our \gls{nmpc} to choose robot orientation adaptively to  any terrain 
is to employ the concept of \emph{mobility}~\cite{focchi17iros}. 
In order to rigorously discuss mobility in mathematical terms, we first define it in words as the attitude of a manipulator (leg) to arbitrarily change end-effector position/orientation~\cite{sciavicco00}.

%
In order to penalize low leg mobility  
in the cost function \eqref{eq:mobility_cost} we need to compute the reference value of hip-to-foot distance ${}_\mathcal{C}\vect{p}_{\mathrm{hf}_{k}}^{\mathrm{ref}}$. Our goal in this section is to define a convenient metric to represent mobility and 
compute ${}_\mathcal{C}\vect{p}_{\mathrm{hf}_{k}}^{\mathrm{ref}}$ corresponding to the maximum value of such 
a metric. 
Among several ways to compute mobility \cite{focchi17iros},  the velocity transformation ratio \cite{Yoshikawa1984} 
allows one to evaluate mobility in a particular direction.
However, the velocity transformation ratio cannot be used in our setting
because it requires prior knowledge of the evolution of the relative 
foot position with respect to \gls{com}. In our case it is not available in advance because it is an output the \gls{nmpc}.  

\begin{figure}
	\centering
	\includegraphics[width=\columnwidth]{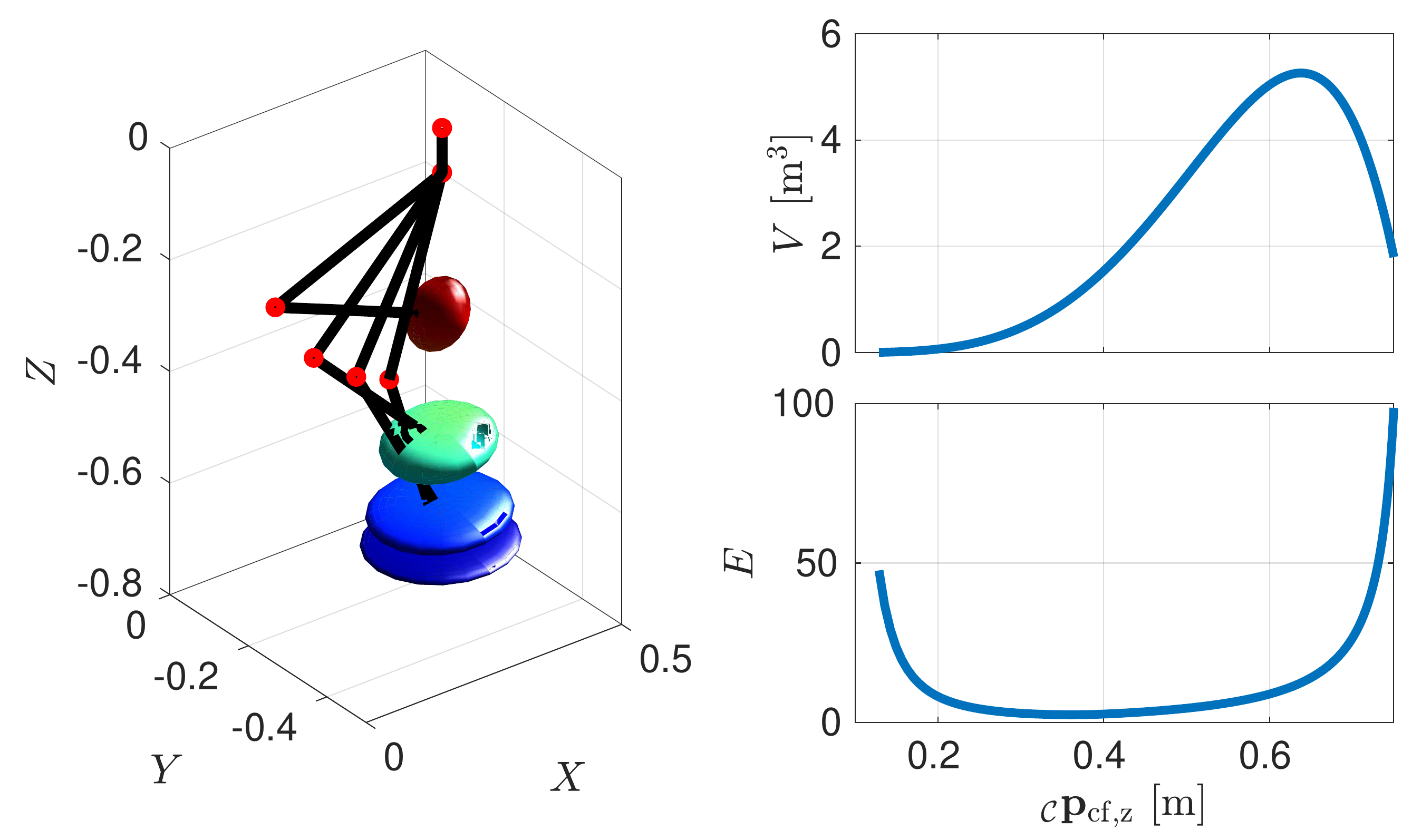}
	\caption{ Manipulability ellipsoid changing with 
		leg configuration (left) of the right front leg. Volume of the ellipsoid (top right) and
		Eccentricity of the ellipsoid (bottom right).}
	\label{fig:3Dellipsoids}
\end{figure}

As an alternative approach, we consider the volume of the \textit{manipulability 
	ellipsoid} $\left( \vect{v} (\mx{J}\mx{J}^\top)^{-1}\vect{v}=1 \right)$ \cite{sciavicco00} as a metric to evaluate mobility. 
A change in the volume of the manipulability ellipsoid with different 
leg configurations is visualized in Fig. \ref{fig:3Dellipsoids} (left).
Inspecting Fig. \ref{fig:3Dellipsoids} (top right), it can be seen that 
the maximum volume $V$ is in the vicinity of the most extended 
leg configuration because the mobility becomes very big in the $X$ and $Y$ direction, 
even if it is still very limited in the $Z$ direction. However, because it is desirable 
to achieve a good mobility in all the directions of the leg configuration, a better metric to do so is the one that accounts for 
the \textit{isotropy} of the manipulability ellipsoid. A measure of the isotropy of an 
ellipsoid can be expressed as the inverse of its \textit{eccentricity} $E$.
Hence, a new  manipulability index that we call \textit{mobility factor} 
\eqref{eq:mobility_factor} can be defined in terms of both the eccentricity 
and the volume of manipulability ellipsoid. 
Again from Fig. \ref{fig:3Dellipsoids}, it can be visualized that 
to keep a good mobility (left plot) in all directions, the volume (top right plot) should be maximized 
while the eccentricity (bottom right plot) as small as possible.
Defining a foot Jacobian $\mx{J}(\vect{q}) \in \Rnum^{3\times3}$ computed at a particular joint configuration $\vect{q}$, the volume $V$ of a manipulability ellipsoid is evaluated as a product 
of the eigenvalues of $\left(\mx{J}(\vect{q})\mx{J}(\vect{q})^\top\right)^{-1}$ 
while the eccentricity is the ratio between its maximum and minimum eigenvalue 
\cite{focchi17iros}.   First, the volume and eccentricity of manipulability ellipsoid
are normalized by their ranges $\bar{V}$ and $\bar{E}$. Then we define the mobility factor as:
\begin{equation}\label{eq:mobility_factor}
m_\mathrm{f} =  \beta \frac{V}{\bar{V}} - \gamma \frac{E}{\bar{E}}
\end{equation}
The minus sign in \eqref{eq:mobility_factor} represents conflicting contributions
of the $V$ and $E$ in the definition of the mobility factor (i.e. the goal is to achieve high volume and low eccentricity). Parameters 
$\beta$ and $\gamma$ are introduced to find a best trade-off between volume and eccentricity 
while deciding a mobility factor.

\begin{figure}
	\centering
	\includegraphics[width=\columnwidth]{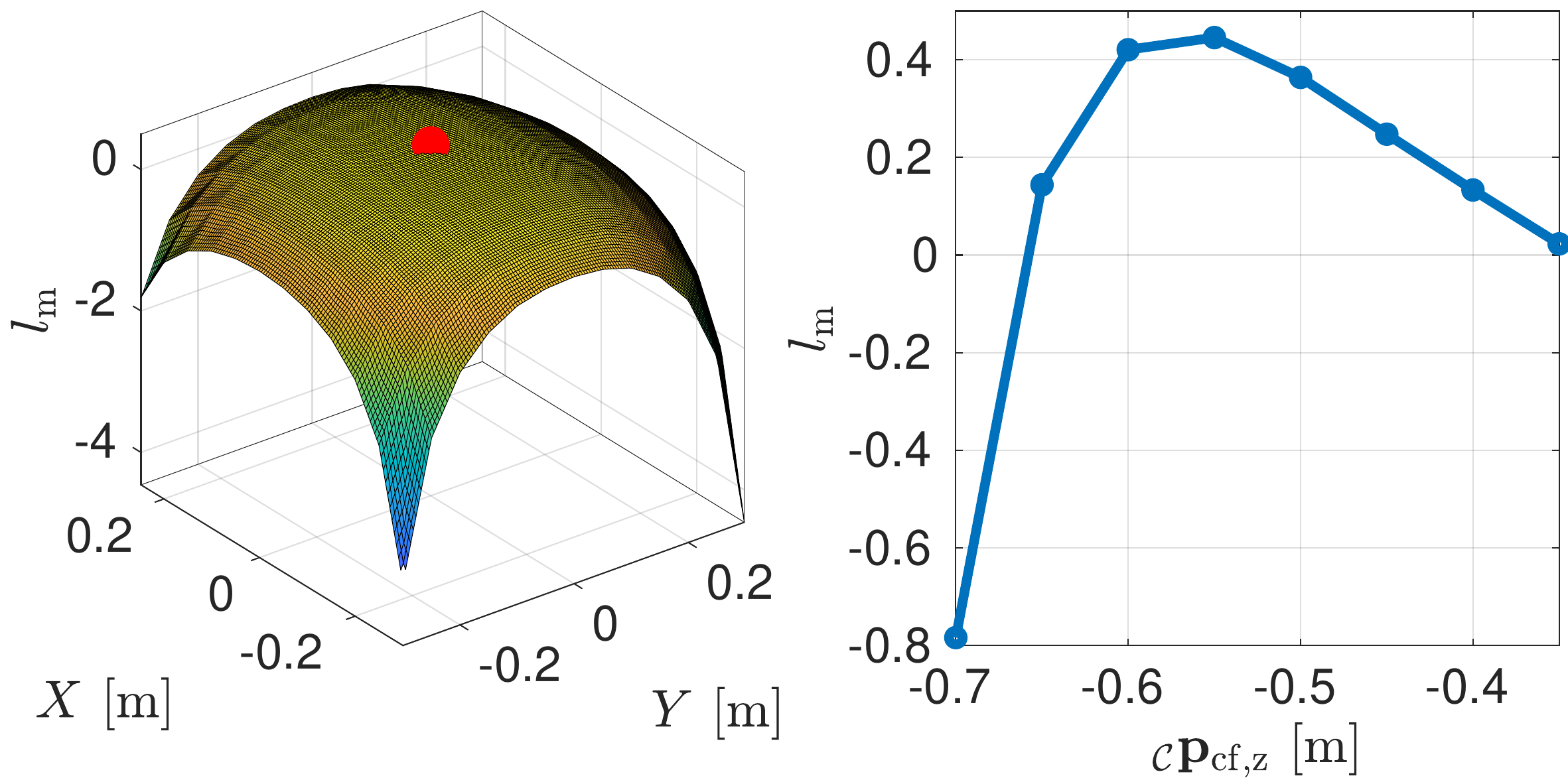}
	\caption{ Slices of the mobility factor function for the \gls{rf} leg: the left figure 
		plots it  against the $X$-$Y$ components keeping $Z$ constant. 
		The red dot in the left plot represents the maxima.
		The right figure plots it against the $Z$ component 
		for a constant $X$-$Y$ foot position. }
	\label{fig:mobility3D}
\end{figure}

The mobility factor is a convex nonlinear function $m_\mathrm{f}: \Rnum^3\to \Rnum$
that can be numerically evaluated inside the workspace of each leg.
By selecting 
$\beta = 1$ and $\gamma = 4$, and after conducting a numerical analysis for all the feet positions in the 
workspace of a leg of the HyQ robot we found that hip-to-foot distance ${}_\mathcal{C}\vect{p}_\mathrm{hf} = (0, 0 , -0.55)\,\mathrm{m}$ 
maximizes $m_\mathrm{f}$. 
In  Fig. \ref{fig:mobility3D} (left) we show a slice of the scalar 
function $m_\mathrm{f}$ in the $X$-$Y$ plane for  ${}_\mathcal{C}\vect{p}_\mathrm{hf,z}=-0.55\,\mathrm{m}$ 
obtained for the \gls{rf} leg. Instead, in  Fig. \ref{fig:mobility3D} (right) 
we plot $m_\mathrm{f}$ against the change of foot position in the $Z$ direction 
considering the hip under the foot ($X=0$, $Y=0$) which clearly 
highlights ${}_\mathcal{C}\vect{p}_\mathrm{hf,z}=-0.55\,\mathrm{m}$ corresponding to
the maximum value of the mobility factor $m_\mathrm{f}$ (i.e., around 0.41). 
We use the output of  this analysis as a reference 
for the hip-to-foot distance
${}_\mathcal{C}\vect{p}_{\mathrm{hf}}^{\mathrm{ref}}$ 
in the mobility cost \eqref{eq:mobility_cost} for all the legs.

In the mobility cost, we multiply $\delta_i$ to the term corresponding to the $i^\mathrm{th}$ leg. Thus, the mobility cost solely accounts for stance legs because the robot can only use them to control its base orientation.
Since, including the mobility cost enables \gls{nmpc} to provide 
the optimal base orientation for a particular locomotion that retains mobility, there is no need to separately specify tracking cost for roll and pitch in the \gls{nmpc}.
This relieves a user from the burden of implementing a customized heuristic (e.g., to align the robot base to the terrain), 
as was necessary in, e.g., \cite{Focchi2020, Gehring2015, Villarreal2020}.
The relative tracking task for the \gls{com} $Z$ position is no longer required either,
because maximizing the mobility in the $Z$ direction automatically takes care of 
keeping an average distance of hips from the terrain to ${}_\mathcal{C}\vect{p}_\mathrm{hf,z}$, 
consequently keeping the robot base at a certain height.

Moreover, the yaw motion results by penalizing the mobility cost along the $X$-$Y$ directions. 
This has the effect of driving the hips of the robot base over the feet, 
naturally aligning the base to the feet, similar to what was 
done in \cite{frontiers20raiola}. However, a tracking cost on yaw was still necessary
in the NMPC to track the  
heading velocity $\omega^{\mathrm{usr}}_\mathrm{z}$ commanded by the user and to avoid oscillations.

\emph{Remark}: The concept of mobility is model independent hence, 
it can also be used with other models such as full body dynamics in the MPC setting.

\subsection{ZMP Margin} \label{subsec:zmp_margin}
In legged locomotion, the robot is often operated close to unstable configurations which require a controller to continuously compensate for model inaccuracies and external disturbances while maintaining locomotion stability. However, a configuration in which the \gls{zmp} \cite{Wieber2009} is close to the boundary of the support polygon \cite{Bretl2008a} could cause instability even with small perturbations due to the loss of control authority. 

In our case, the reference generator computes  
references for the \gls{grfs} by dividing the robot mass with the 
number of legs as explained in Section \ref{sec:ref_gen}. 
Penalizing \gls{grfs} $Z$ component heavily in the tracking cost \eqref{eq:tracking_cost} ensures that they stay close to the reference, consequently maintaining a higher loading on the diagonally opposite leg to the swinging one, and therefore maintaining some margin for the locomotion stability.
To evaluate the locomotion stability, we define the \emph{\gls{zmp} margin} which is computed as the minimum of the distance of the \gls{zmp} from each support polygon edge, i.e.,
\begin{equation}
m_\mathrm{c} = \min(\vect{d})
\end{equation}
where $\vect{d}$ is a vector of the distances of \gls{zmp} projection (on a horizontal plane) from the support polygon edges.

%
%


\subsection{Force Robustness} \label{subsec:force_robustness}
Similar to the considerations on mobility, in order to  effectively compensate for disturbances acting on the system, robustness in the \gls{grfs} is required. The closer the GRF is to the friction cone boundary, the less lateral force is available to compensate for perturbations. 
An approach penalizing \gls{grfs} that are in the vicinity of the cone boundaries has been proposed in \cite{Focchi2016, Fahmi2020TRO} inside the \gls{wbc}. These \gls{wbc} based approaches instantaneously generate \gls{grfs} that are as close as possible to the normals of the cones while yielding the prescribed resultant wrench on the robot base. However, \gls{wbc} does not account for the future state of the robot and hence, it leaves some room for the \gls{nmpc} to compensate for the contact normal estimation error and recover from external disturbances. 
Introducing these margins on \gls{grfs} from the cone boundaries is especially important in some scenarios, such as the one reported in simulation in Section \ref{sim:vshaped_wall}. 

In this paper, we adopt a similar idea to \cite{Focchi2016, Fahmi2020TRO} and introduce the additional cost term \eqref{eq:force_robustness} in the \gls{nmpc}, which penalizes the tangential components of \gls{grfs} in the contact frame $\mathcal{K}$ (see Fig. \ref{fig:hyqScheme}) to obtain the resultant \gls{grfs}  as close as possible to the contact normals. The weight matrix $\mx{P}$ used in this cost is defined in Table \ref{tab:nmpc_weights} (Section \ref{sec:results}). Note that it is required to penalize the $X$-$Y$ components higher than $Z$ component of \gls{grfs} in the contact frame to achieve this behaviour.
\section{Reference Generator}
\label{sec:ref_gen}
In our approach, the \gls{nmpc} requires a reference trajectory of 
the state and control input along with the model parameters i.e., 
foot positions and contact status. 
For the very first run of the NMPC, this reference trajectory also serves  as an initial guess.
The references are generated for the length of control intervals $N$, 
since the reference generator is called before every iteration of
the NMPC in order to obtain prompt adaptation to terrain changes and user set-point.
Our reference generator is based on heuristics and it takes as inputs: 
\begin{itemize}
	\item the user commanded longitudinal and lateral \gls{com} velocity  
	${}_\mathcal{H} \vect{v}_\mathrm{c}^{\mathrm{usr}} \in \mathbb{R}^2$ in the 
	horizontal frame $\mathcal{H}$,\footnote{The horizontal frame is placed like the \gls{com} frame
		but with the  $Z$-axis aligned with the gravity}
	\item user commanded heading velocity $\omega^{\mathrm{usr}}_\mathrm{z} \in \Rnum$,
	\item current pose of the robot $(\vect{p}_\mathrm{c}, \bm{\Phi})$,
	\item current feet positions  $\vect{p}_\mathrm{f} \in \Rnum^{12}$,
	\item heightmap of the terrain
\end{itemize} 
The reference generator outputs:
\begin{itemize}
	\item the references for the \gls{nmpc} cost: states 
	$\vect{x}^\mathrm{ref} \in  \Rnum^{n_x \times (N+1) }$,
	control $\vect{u}^\mathrm{ref} \in  \Rnum^{n_u \times N }$,
	\item parameters $\vect{a}$ of the model: sequence of the contact status $\bm{\delta}$ ($ \in  \Rnum^ {4 \times N}$) 
	and sequence of the foot locations $\vect{p}_\mathrm{f}$ ($\in  \Rnum^{12 \times N}$),
	\item normals of the terrain at the foothold locations, which are
	provided as inputs to the \gls{nmpc} for the cone constraints.
\end{itemize} 
%
%

First we compute the $X$-$Y$ components of the total velocity 
$\vect{v}^\mathrm{ref}_\mathrm{c} \in \mathbb{R}^3$, which depend on both $\vect{v}_\mathrm{c}^{\mathrm{usr}}$ and 
the $X$-$Y$ components of the tangential velocity due 
to the heading velocity $\bm{\omega}^{\mathrm{usr}}$ = $(0, 0, \omega^{\mathrm{usr}}_\mathrm{z})$.
\begin{equation}
\vect{v}^\mathrm{ref}_{\mathrm{c,(x,y)}} = \vect{v}_\mathrm{c}^{\mathrm{usr}} + (\bm{\omega}^{\mathrm{usr}} \times \vect{p}_\mathrm{c}^\mathrm{ref})_\mathrm{(x,y)}
\end{equation}
The $X$-$Y$ \gls{com} position $\vect{p}^\mathrm{ref}_\mathrm{c,(x,y)}$ is obtained by 
integrating the $\vect{v}^\mathrm{ref}_\mathrm{c,(x,y)}$ with  the explicit Euler scheme. 
The references for \gls{com} $Z$, roll and pitch are set to $0$ 
because we do not track them in the \gls{nmpc} cost \eqref{eq:tracking_cost}.
Instead, the reference for the yaw $\psi$ is obtained  by integrating the user defined 
yaw rate $\dot{\psi}^\mathrm{usr}$ with $\dot{\bm{\Phi}}^{\mathrm{usr}} = \vect{E}^{-1}
(\bm{\Phi}^{\mathrm{ref}}) \bm{\omega}^{\mathrm{usr}}$ (see Appendix \ref{adx:eulerRate} 
for the transformation between angular velocity and Euler rates). 
The reference for angular velocity, instead, coincides with $\bm{\omega}^{\mathrm{usr}}$.\\
%
%
%
%
%
%
%
%
%
The references for \gls{grfs} $\vect{u}^\mathrm{ref}$ are calculated by simply dividing the 
total mass of the robot by the number of legs in stance.
Dividing the forces equally onto the legs is correct only if the robot is static, 
but, in case of dynamic conditions, it is a better approximation than passing no references.

The sequence of contact status $\bm{\delta}$ and of footholds are computed by the 
\textit{gait scheduler} and \textit{robocentric stepping} strategy, respectively.
It is important to mention that the reference generator does
not compute the swing trajectories and they are obtained from the
\gls{wbc} interface discussed in Section \ref{subsec:wbc_interface}. 

\subsubsection{Gait scheduler}
\begin{figure}[!tb]
	\centering
	\includegraphics[width=0.68\columnwidth]{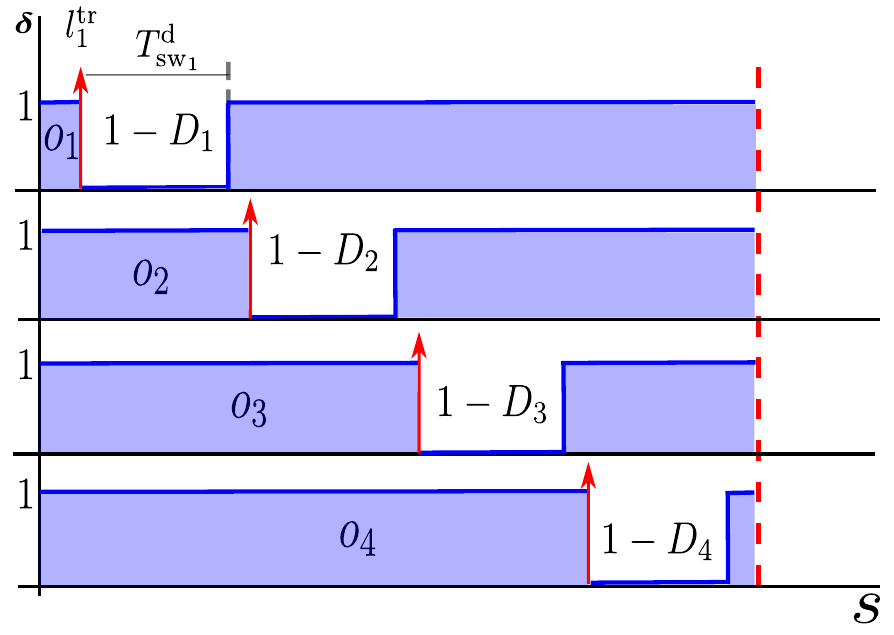}  
	\includegraphics[width=0.3\columnwidth]{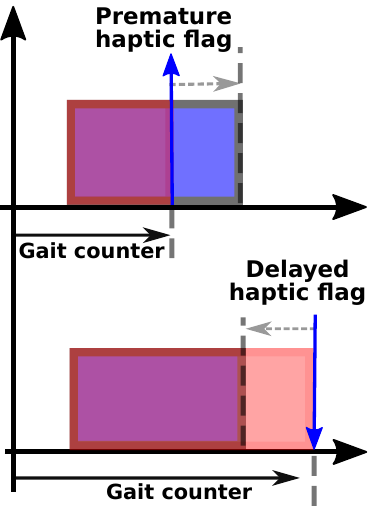} 
	\caption{Gait schedule for a walk. Offsets $\vect{o} =\left[0.05, 0.3, 0.55,0.8\right]$, 
		duty-factors $\vect{D} =\left[0.85, 0.85, 0.85,0.85\right]$.
		The red arrows represent the trigger $l^\mathrm{tr}_{i}$ for a swing leg $i$. 
		Right part shows the fast-forwarding (top) or re-winding  (bottom)
		of the gait counter to recover synchronization between actual (haptic) and planned touchdown.}
	\label{fig:gaitScheduler}
\end{figure}

The gait scheduler is logically decoupled from the reference trajectory generation and 
determines if a leg is either in swing or in stance $({\delta}_i)$ at 
each time instance for the entire gait cycle as shown in Fig. \ref{fig:gaitScheduler} (left).

The leg duty factor $D_i$ and offsets $o_i$ can be used to encode 
different gaits such as crawl, trot and pace. The gait scheduler implements
a time parametrization $s \in [0,1]$ (\textit{stride phase}) which is normalized 
about the cycle time duration $T_\mathrm{c}$ such that the leg duty 
factor $D_i$ and offsets $o_i$ are independent from the cycle time. 
Each trigger $l^\mathrm{tr}_i$ (red flag in Fig. \ref{fig:gaitScheduler} (left)) 
corresponds to a new lift-off event.
%
%
%
We can express the value of $\delta$ for leg $i$ as:
\begin{equation}
\delta_i =\begin{cases}
1 , &   s < o_i  \vee s >((o_i + (1-D_i)) \bmod 1)\\
0, & \text{otherwise}
\end{cases}
\end{equation}
%
%
%
%

Every time the reference generator is called, it extracts $N$ 
points from the gait schedule starting from an index called \textit{gait counter}. 
It keeps memory of the index of the gait schedule achieved by the previous call of the reference generator.
The synchronization between the first point of a contact sequence $\delta_{i_k}$ 
computed by the reference generator and the actual contact state of the robot 
avoids the reference generator to compute a zero reference force while the leg is in stance and vice-versa.
In the case of premature or delayed touchdown events, the synchronization is lost and
the gait counter is shifted backwards or forward
to re-conciliate the planned touchdown with the actual 
touchdown as shown in Fig.  \ref{fig:gaitScheduler} (right).
This is a crucial feature when dealing with rough terrains.
%
%

\subsubsection{Robocentric stepping}
\label{sec:robocentric_stepping}
\begin{figure}[!tb]
	\centering 
	\includegraphics[scale=0.195]{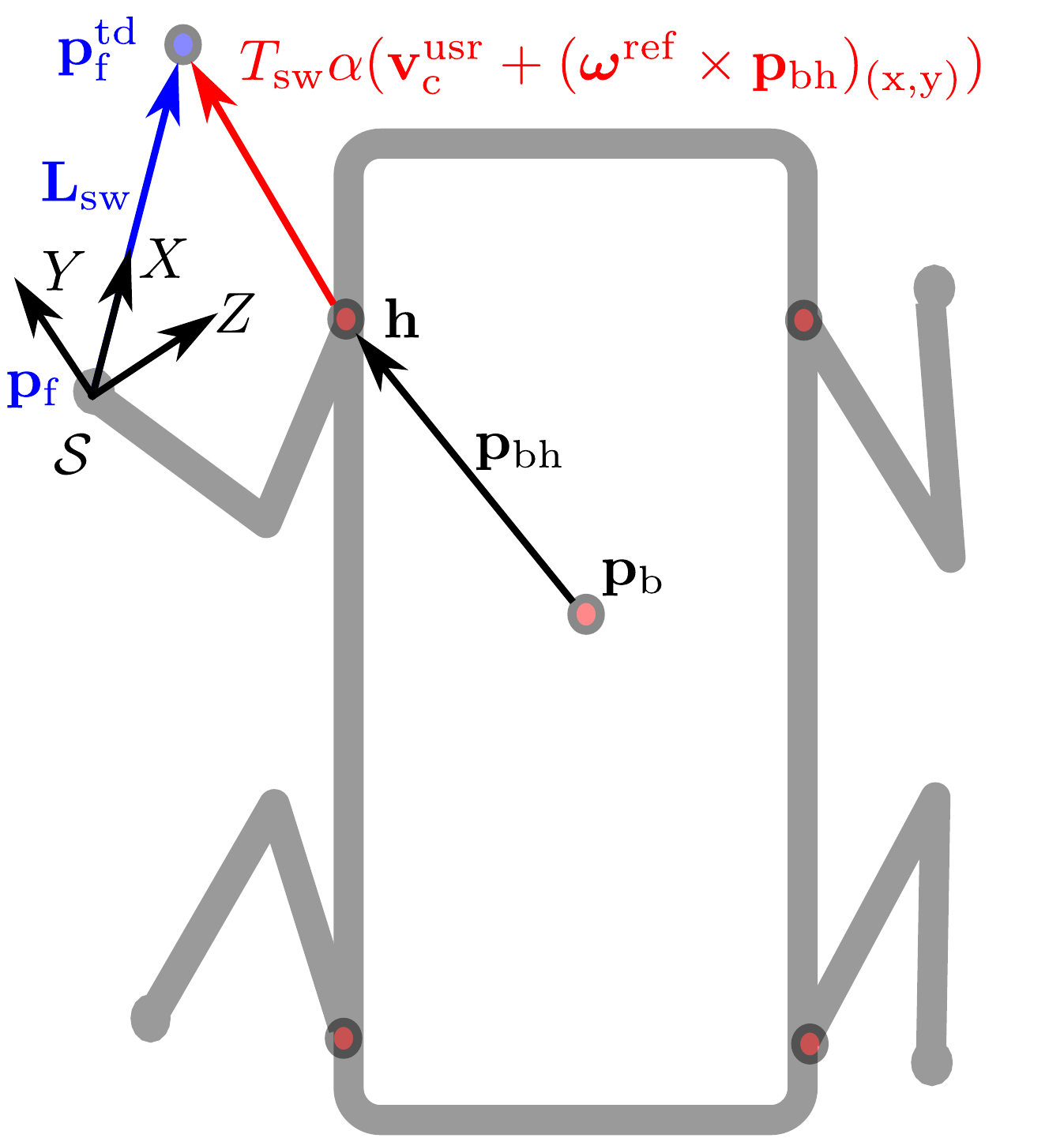}  
	\caption{ Representation of the robocentric stepping strategy and of the Swing Frame,
		located at the lift-off point. The red arrow shows the distance between the 
		touchdown point $\vect{p}^\mathrm{td}_\mathrm{f}$ and the hip $\mathbf{h}$. 
		The blue vector $\vect{L}_\mathrm{sw}$ connects lift-off and touchdown point.}
	\label{fig:robocentricStepping}
\end{figure}

%
The choice of the foothold is a key element in locomotion, 
since it deals with the kinematic limits of the robot. 
Inspired by \cite{raibert1986legged}, we use an approach that continuously computes  
footholds consistent with the current position of the robot.  
To compute a foothold for a swinging leg $i$, we
consider its hip position  $\vect{h}_i$ instead of 
using the foot position at the moment of  lift-off.
In this way a disturbance acting on the robot or a tracking error 
occurred during a swing  can be recovered in the following swing. 
For a leg $i$, dropping the index to simplify the notation and defining 
the lift-off trigger as $l^\mathrm{tr}_k=  \delta_k  \land \overline{\delta}_{k+1}$, the foot position is computed as:
\begin{equation} \label{eq:footposition}
\vect{p}_{\mathrm{f}_{k+1}}= 
\begin{cases}
\vect{p}^{\mathrm{td}}_{\mathrm{f}_k} & l^\mathrm{tr}_k= 1 \ \\
\vect{p}_{\mathrm{f}_k} & l^\mathrm{tr}_k= 0
\end{cases}
\end{equation}
Notice that at the lift-off condition  $l^\mathrm{tr}= 1$ at instance $k$, $\vect{p}_\mathrm{f}$ is set equal 
to the touchdown point  $\vect{p}^{\mathrm{td}}_\mathrm{f}$ and it is kept constant until the
next lift-off event occurs. The $X$-$Y$ component of the touchdown point is given by:
\begin{equation} \label{eq:touchdownpoint}
\vect{p}^\mathrm{td}_{\mathrm{f}_k,\mathrm{(x,y)}} = 
\vect{h}_k +  \alpha T_\mathrm{sw}^\mathrm{d}  (\vect{v}_\mathrm{c}^{\mathrm{usr}}  + (\bm{\omega}^{\mathrm{usr}} \times \vect{p}_\mathrm{bh})_\mathrm{(x,y)}   )
\end{equation}
%
%
The second term in \eqref{eq:touchdownpoint} represents the step 
length (red arrow in Fig.~\ref{fig:robocentricStepping}) which is computed 
with respect to the hip instead of the previous foot location.
Parameter $\alpha$ is an empirically chosen scaling factor. 
Parameter $T_{\mathrm{sw}}^\mathrm{d}$ is the default swing duration computed starting 
from user-defined offsets $\vect{o}$ and duty-factors $\vect{D}$.
$\vect{p}_\mathrm{bh}\in \Rnum^3$ is the distance between hip and center of the base.
%
%
%
%
A 2.5D heightmap of the terrain  is evaluated in correspondence 
of the touchdown point $\vect{p}^\mathrm{td}_{\mathrm{f}_k,\mathrm{(x,y)}}$, 
to obtain $\vect{p}_{\mathrm{f}_k,\mathrm{z}}^\mathrm{td}$ that does not penetrate the terrain.
If $\vect{p}_\mathrm{f}^{\mathrm{td}}$ is located near an edge or 
leads to collisions (e.g., of the foot or the shin) during the step cycle, 
this can be harmful for the robot's balance.
To prevent this from happening, the robot acquires a local heightmap 
in the vicinity of the touchdown point $\vect{p}_\mathrm{f}^{\mathrm{td}}$ 
and adjusts the foot landing location using the \gls{vfa} module presented in \cite{villarreal19ral}.
\section{Whole-Body Controller} \label{sec:whole_body_controller}
In this section, we describe the \gls{wbc} that tracks planned trajectories
$\vect{x}^\mathrm{p}$ and $\vect{u}^\mathrm{p}$ provided by the \gls{nmpc}. 
The \gls{wbc} first computes feed-forward $\mx{W}^\mathrm{d}_\mathrm{ff}$ 
and feedback $\mx{W}^\mathrm{d}_\mathrm{fb}$ wrenches from the planned 
trajectories and then the sum of these wrenches are mapped into \gls{grfs} 
through the \gls{qp} optimization \eqref{eq:qp_wbc}. The \gls{wbc} also
maps the \gls{grfs} 
into the joint torques $\bm{\tau}^*$. This joint torque along with 
low-impedance feedback torque $\bm{\tau}_\mathrm{fb}$ results in 
the total torque $\bm{\tau}_\mathrm{d}$ required by the low-level joint torque control block. 
Refer to Fig.~\ref{fig:blockDiagram} for the block representation of \gls{wbc} inside our locomotion framework.
%
\subsection{WBC interface}\label{subsec:wbc_interface}
In our planning framework, the \gls{nmpc} runs at re-planning frequency of $25\,\mathrm{Hz}$  whereas the \gls{wbc} 
requires  state and control inputs at  $250\,\mathrm{Hz}$ 
(we will call this the \textit{\gls{wbc}  frequency}). 
Hence, we introduce a \gls{wbc} interface block that re-samples state 
and control inputs at the \gls{wbc} frequency. 
In particular, in order to obtain the desired $\vect{u}^\mathrm{d}$ we use a zero-order hold filter of $\vect{u}^\mathrm{p}$. 
The planned states $\vect{x}^\mathrm{p}$ from the \gls{nmpc}, instead,
are re-sampled with a linear interpolation to obtain $\vect{x}^\mathrm{d}$.
\footnote{The rigorous approach is to use the model \eqref{eq:newtonEuler} 
	to predict the evolution of the system in the $T_\mathrm{s}$ time interval, 
	considering the $\vect{u}^\mathrm{p}$ coming from the \gls{nmpc}, 
	but for the motions considered in this paper the result is very similar, 
	so a linear interpolation is a fair approximation.}

Finally, the feed-forward wrench $\mx{W}_\mathrm{ff}\in \Rnum^6$ is computed  
from the desired \gls{grfs} $\vect{u}^\mathrm{d}$ as:
\begin{equation}\label{eq:ffwd_wrench}
\mx{W}^\mathrm{d}_\mathrm{ff}= \mat{\sum_{i=1}^4 \vect{u}^{\mathrm{d}}_i &   
	\sum_{i=1}^4 \vect{p}_{\mathrm{cf},i}^\mathrm{d} \times \vect{u}^\mathrm{d}_i}^\top
\end{equation}
%
%


\subsection{Feedback wrench}\label{feedbackController}
We use the approach of \cite{Focchi2016} 
to define desired feedback wrench obtained from a Cartesian impedance 
and briefly recall it next for completeness:
%
%
\begin{equation}
\mx{W}^\mathrm{d}_\mathrm{fb}=\mx{K}
\mat{\vect{p}_\mathrm{c}^{\mathrm{d}}-\vect{p}_\mathrm{c}\\
	\mx{e}({}_{\mathrm{w}}\mx{R}_\mathrm{b}^{\mathrm{\top}} {}_{\mathrm{w}}\mx{R}_\mathrm{d})} 
+\mx{D}
\mat{\mx{v}^{\mathrm{d}}_\mathrm{c}-\mx{v}_\mathrm{c}\\
	\bm{\omega}^\mathrm{d}_\mathrm{b}-\bm{\omega}_\mathrm{b}}
\end{equation}
%
%
%
where
${}_{\mathrm{w}} \mx{R}_\mathrm{b}$ and $ {}_{\mathrm{w}} \mx{R}_\mathrm{d} \in \Rnum^{3\times3} $ 
are the rotation matrices representing actual and desired orientation of the
base with respect to the inertial frame, respectively, $\mx{e}(\cdot) : \Rnum^{3\times3} \to \Rnum^{3}$ is
a mapping from a rotation matrix to the associated rotation vector.
Matrices $\mx{K}$ and $\mx{D}$ 
are diagonal matrices containing the proportional and derivative gains and they can be interpreted as impedances. 

\textit{Remark:} At each re-planning instance of \gls{nmpc}, 
the state reference is computed from the current state of the robot $\vect{\hat{x}_0}$. 
Thus,  at each  re-planning instance the feedback term is nullified.
%
%
\subsection{Projection of the GRF\lowercase{s}}

While the feedforward wrenches $\mx{W}^\mathrm{d}_\mathrm{ff}$  provided by MPC 
satisfy the friction cone and unilateral constraints by construction, this guarantee 
is lost with the addition of the feedback term $\mx{W}^\mathrm{d}_\mathrm{fb}$  to the wrenches. 
Therefore, one needs to project the total wrenches $\mx{W}^\mathrm{d}_\mathrm{ff}+\mx{W}^\mathrm{d}_\mathrm{fb}$
onto the set of wrenches that satisfy the constraints.
The matrix representation 
\begin{equation}\label{eq:linear_sys_wbc}
\underbrace{\mat{\delta_1 \mx{I} & \ldots & \delta_4\mx{I} \\ 
		\delta_1\left[\vect{p}_{\mathrm{cf}, 1} \times\right] & \ldots & \delta_4\left[\vect{p}_{\mathrm{cf}, 4} \times\right]}}_{\mx{A}}
\underbrace{\left[\begin{array}{c}\vect{f}_{1} \\ \vdots \\ \vect{f}_{4}\end{array}\right]}_{\vect{f}}
=\underbrace{\mx{W}^\mathrm{d}_\mathrm{ff}+\mx{W}^\mathrm{d}_\mathrm{fb}}_{\vect{b}}
\end{equation}
is derived from a 
simplified \gls{srbd} model \cite{Focchi2016} and allows us to map the 
desired wrenches into \gls{grfs}.
%
%
%
%
To compute the desired \gls{grfs} $\vect{f}^{\mathrm{d}}$ 
we solve the following \gls{qp}:
\begin{subequations} \label{eq:qp_wbc}
	\begin{align} 
	\mx{f}^\mathrm{d} =  \argmin_{\vect{f}} \ & \parallel \mx{A} \vect{f}-\vect{b}\parallel_{\mx{S}}^2 +
	\parallel \vect{f}-\vect{u^\mathrm{d}}\parallel_{\mx{T}}^2 \\ 
	\text{s.t.} \quad &  	\underline{\vect{d}}\leq \vect{C}\vect{f}\leq \overline{\vect{d}} \label{eq:wbc_qp_constraints}
	\end{align}
\end{subequations}
%
%
The term $ \parallel \vect{f}-\vect{u^\mathrm{d}}\parallel_{\mx{T}}^2$ in the cost \eqref{eq:qp_wbc} 
allows the tracking of the desired forces $\vect{u}^\mathrm{d}$ received from the \gls{nmpc}. 
Matrices
$\mx{S} \in \Spd{6}$ and $\mx{T} \in \Spd{12}$ are positive-definite weight matrices. 
Inequality \eqref{eq:wbc_qp_constraints} encodes the friction cone and 
unilateral constraints similar to \eqref{eq:friction_cone} for which
further details can be found in \cite{Focchi2016}. 
It is important to note that gravity compensation is already incorporated in the \gls{nmpc} 
formulation through the \gls{srbd} model.
\subsection{Mapping GRF\lowercase{s} to Joint Torques}
%
The \gls{grfs}  $\mx{f}^\mathrm{d}$ must be mapped into joint torques
$\bm{\tau}^*$. We do so by exploiting the joint dynamics:
\begin{equation}
\bm{\tau}^* = -\mx{J}(\vect{q}) ^\top \vect{f}^\mathrm{d} +
\vect{h}(\vect{q},\dot{\vect{q}})
\end{equation}
where $\vect{J}(\vect{q}) \in \Rnum ^{n_u \times n_x}$ is the contact Jacobian  
and $\vect{h}(\vect{q},\dot{\vect{q}})$ the vector of gravity/Coriolis terms in the leg joint dynamics. 
We neglect the joint acceleration contribution, because 
it is very small with respect to the other terms.
\subsection{Joint-Space PD}
A $1 \,\mathrm{kHz}$ Joint-Space PD is put in cascade with the \gls{wbc} 
before sending torques to the low-level controller. In this way, we track the desired
trajectories of the swinging legs and we increase the robustness in case a foot loses contact with the ground.
The \gls{wbc} interface provides the 
joint trajectories $\vect{q}^\mathrm{d}$ and $\dot{\vect{q}}^\mathrm{d}$ required by the Joint-Space PD.
To compute the joint trajectories, inverse kinematics is required which in turn 
needs the swing trajectory $\vect{p}^{\mathrm{sw}}_\mathrm{f}$.
We define the swing frame $\mathcal{S}$ \cite{frontiers20raiola} (Fig. \ref{fig:robocentricStepping}), 
whose $X$-axis is aligned with the vector that links lift-off and
touchdown point ($\vect{L}_\mathrm{sw}$), $Y$-axis is perpendicular to the $X$-axis
of the swing frame and to the $Z$-axis of the world frame.
Finally the $Z$-axis is such that $\mathcal{S}$ is a  counter-clockwise coordinate system. 
The origin of the swing frame $\mathcal{S}$ coincides with the 
lift-off point. 
In this way the swing trajectory lies on the $X$-$Z$ plane and we shape it as a semi-ellipse 
with $L_\mathrm{sw}$ and $H_\mathrm{sw}$ as lengths of the axes:
\begin{align}
{}_\mathcal{S} \vect{p}^{\mathrm{sw}}_f = \mat{
	\frac{L_\mathrm{sw}}{2} (1- \mathrm{cos}(\pi f_\mathrm{sw} t_\mathrm{sw}))\\
	0.0\\
	H_\mathrm{sw} \mathrm{sin}(\pi f_\mathrm{sw} t_\mathrm{sw})}
\end{align}
where $t_\mathrm{sw}$ is the time elapsed from the beginning of a swing 
and $f_\mathrm{s} = 1/T_{\mathrm{sw}}^\mathrm{d}$ is the swing frequency. 
We map ${}_\mathcal{S}\vect{p}^{\mathrm{sw}}_\mathrm{f}$and its derivative 
in the inertial frame $\mathcal{W}$ to obtain $\vect{p}^{\mathrm{sw}}_\mathrm{f}$ and $\dot{\vect{p}}^{\mathrm{sw}}_\mathrm{f}$, respectively.  Finally, after  evaluating the relative 
foot position ${}_\mathcal{C}\vect{p}_\mathrm{cf}$ and velocity ${}_\mathcal{C}\dot{\vect{p}}_\mathrm{cf}$
we can obtain $\vect{q}^\mathrm{d}$ and $\dot{\vect{q}}^\mathrm{d}$ via inverse kinematics.\\
\section{\gls{rti} for \gls{nmpc}} \label{sec:rti_scheme}
One of the main drawbacks of \gls{nmpc} is its computational burden, thus 
efficient tailored algorithms are necessary in order to achieve fast sampling
rates for complex systems with fast dynamics.
While many approaches have been developed for optimal control, a complete
discussion about all possible approaches is beyond the scope of this paper. 
We focus on direct multiple shooting methods derived from \gls{sqp} 
that have been specifically developed for real-time \gls{nmpc}~\cite{Diehl2009}.

In multiple shooting methods both state $\vect{x}$ and control input $\vect{u}$ are 
decision variables unlike in single shooting where the decision vector only includes 
the control input. We ought to stress that this does not increase the computational 
complexity with respect to single shooting (where computations are moved from linear 
algebra to the evaluation of derivatives).
Furthermore, multiple-shooting allows one to provide an initial guess also for the
state trajectory, which is typically beneficial for unstable systems in an \gls{nmpc} context~\cite{Diehl2009}.


\gls{sqp} is a  popular algorithm which
solves an \gls{nlp} by iteratively solving local quadratic approximations (\gls{qp}s) of the problem~\cite{Gros2020}. 
At each \gls{sqp} iteration, the solution from the previous step is recycled to define an initial guess 
($\vect{x}_k^{\mathrm{L}}, \vect{u}_k^{\mathrm{L}}$), which is then used to construct a \gls{qp} approximation of
the \gls{nlp}~\eqref{eq:nlp_formulation}, given by
\begin{subequations}\label{eq:qp_subproblem}
	\begin{align}
	\min _{\Delta \vect{x}, \Delta \vect{u}} \quad & \sum_{k=0}^{N-1} \frac{1}{2}\left
	[\begin{array}{c}\Delta \vect{x}_k \\ \Delta \vect{u}_k\end{array}\right]^\top 
	\mx{H}_k
	\left[
	\begin{array}{c}\Delta \vect{x}_k \\ \Delta \vect{u}_k\end{array}
	\right]+
	\vect{J}_k^\top
	\left[
	\begin{array}{c}\Delta \vect{x}_k \\ \Delta \vect{u}_k\end{array}
	\right] \label{eq:qp_cost_function} \\
	\mathrm{s.t.} \quad	& \Delta \vect{x}_0=\hat{\vect{x}}_0-
	\vect{x}_0^{\mathrm{L}}, \label{eq:qp_initial_condition}\\
	&\Delta \vect{x}_{k+1}=\mx{A}_k \Delta \vect{x}_k+\mx{B}_k \Delta \vect{u}_k+\vect{r}_k, \label{eq:qp_equality}\\
	&\mx{C}_k \Delta \vect{x}_k+\mx{D}_k \Delta \vect{u}_k+\vect{h}_k \geq 0, \label{eq:qp_inequality}
	\end{align}
\end{subequations}
where,
$\Delta \vect{x}_k = \vect{x}_k-\vect{x}^\mathrm{L}_k, \Delta \vect{u}_k = \vect{u}_k-\vect{u}^\mathrm{L}_k$, 
$\hat{\vect{x}}_0$ is the current system state, and
\begin{subequations}\label{eq:qp_sensitivities}
	\begin{align*}
	\mx{A}_k &=\left.\frac{\partial f(\vect{x}, \vect{u}, \vect{a}_k)}{\partial \vect{x}}\right|_{\vect{x}_k^{\mathrm{L}}, \vect{u}_k^{\mathrm{L}}}, 
	&\mx{B}_k &=\left.\frac{\partial f(\vect{x}, \vect{u}, \vect{a}_k)}{\partial \vect{u}}\right|_{\vect{x}_k^{\mathrm{L}}, \vect{u}_k^{\mathrm{L}}}, \\ 
	\mx{C}_k  &=\left.\frac{\partial h(\vect{x}, \vect{u}, \vect{a}_k)}{\partial \vect{x}}\right|_{\vect{x}_k^{\mathrm{L}}, \vect{u}_k^{\mathrm{L}}} ,  
	&\mx{D}_k &=\left.\frac{\partial h(\vect{x}, \vect{u}, \vect{a}_k)}{\partial \vect{u}}\right|_{\vect{x}_k^{\mathrm{L}}, \vect{u}_k^{\mathrm{L}}}, \\
	\vect{r}_k&=g\left(\vect{x}_k^{\mathrm{L}}, \vect{u}_k^{\mathrm{L}}, \vect{a}_k\right)-\vect{x}_{k+1}^{\mathrm{L}}, 
	&\vect{h}_k&=h\left(\vect{x}_k^{\mathrm{L}}, \vect{u}_k^{\mathrm{L}}, \vect{a}_k\right)\\
	\vect{J}_k&=\mx{W}_k\left[\begin{array}{c}\vect{x}_k^{\mathrm{L}}-\vect{x}_k^{\text {ref }} \\ \vect{u}_k^{\mathrm{L}}-\vect{u}_k^{\text {ref }}\end{array}\right] \tag{\ref{eq:qp_sensitivities}}
	\end{align*}
\end{subequations}
Matrix $\mx{H}_k$ is the diagonal blocks of a suitable approximation of the 
Lagrangian Hessian. Since our problem relies on a least-squares cost, we adopt
the popular Gauss-Newton Hessian approximation~\cite{Gros2020} that gives $\mx{H}_k = \mx{W}_k$. 

While in \gls{sqp} one solves several \gls{qp}s until convergence is reached, 
the \gls{rti} scheme consists in solving a single \gls{qp} per sampling time.
This is motivated by the observation that in \gls{nmpc} two subsequent problems 
have very similar solutions. Therefore, by reusing the solution of the previous
\gls{nmpc} problem, one obtains a very good initial guess for the next problem,
which essentially only needs to correct for external perturbations and model mismatch. 
For all details on the \gls{rti} scheme, we refer to~\cite{Diehl2005,Gros2020}
and references therein. We limit ourselves to observe that, 
since ($\vect{x}_k^{\mathrm{L}}, \vect{u}_k^{\mathrm{L}}$) is known \emph{before} 
the next state measurement is available, one can already evaluate the functions and their derivatives~\eqref{eq:qp_sensitivities} before the initial state $\vect{\hat x}_0$ 
is available. Consequently, the \gls{qp} can be constructed and prepared beforehand; 
note that this also includes the first factorization of the \gls{qp} Hessian. 
Once $\vect{\hat x}_0$ is available, one only has to finish solving the \gls{qp}. 
Therefore, while the overall sampling time must still be long enough to prepare
the next \gls{qp}, the latency between the time at which $\vect{\hat x}_0$ 
is available and the time at which the control input can be applied to the system is very small.

Note that in the \gls{rti} scheme proposed above, the functions and their 
derivatives~\eqref{eq:qp_sensitivities} are evaluated along a guess obtained 
from the previous solution, rather than along the reference trajectory. 
Another important aspect to highlight is the fact that there exist several 
approaches to compute~\eqref{eq:qp_sensitivities}. One choice consists of 
first linearizing the continuous-time system dynamics and then using the 
matrix exponential to obtain a discrete-time linear system. This approach 
presents some advantages, but can be computationally demanding. For the time-varying and infeasible references, however, it is preferred to \textit{first discretize and then linearize}~\cite{Gros2020}. In this work we deal with time-varying and infeasible references, hence we opt for first discretize and then linearize approach. An advantage of this apporach is that after numerically approximating the discrete-time dynamics, the linearization can be obtained at a desired accuracy.


A very popular
way to obtain discret-time dynamics is with the explicit Euler
integrator, which is computationally inexpensive, but can be inaccurate and unstable.
Therefore, it is usually more efficient to resort to 
higher-order integration schemes, such as, e.g., the popular Runge-Kutta methods. 
Finally, we should further stress that there also exist implicit integration schemes,
which require more computations per step, but they are typically much more stable
and accurate than explicit schemes for some classes of systems. Unfortunately, 
the selection of the least computationally demanding integrator which delivers 
sufficient accuracy depends on the problem setting and typically requires some 
trial-and error approach, which can be educated using some guidelines based on
the theoretical properties of each integrator~\cite{Quirynen2017,Quirynen2014a,Hairer1993,Hairer1996,Butcher2003}. 

In this work, we relied on the \gls{rti} implementation provided by $\texttt{acados}$~\cite{Verschueren2019},
which consists of tailored efficient implementations of \gls{qp} solvers, 
numerical integration schemes, and all other components of the \gls{rti} scheme.

\section{Results}\label{sec:results}
In this section we discuss the implementation details and results obtained 
from the simulations and experiments with the \gls{nmpc} scheme proposed
in Section \ref{sec:nmpc}.
\subsection{Implementation details} 
\label{sec:implementation}
To check the efficacy of our \gls{rti} based \gls{nmpc} algorithm with the proposed 
features mentioned in Section \ref{sec:features}, we performed 
several simulations and experiments in challenging scenarios.  
The simulation and experiments were performed on the \gls{hyq} robot of mass $m = 87 \,\mathrm{kg}$.
The feedback gains used in the \gls{wbc} are $\mx{K} = \mathrm{diag}(1500,1500,1500, 100,100,100)$ and $\mx{D} = \mathrm{diag}(1000,1000,1000,50,50,50)$. We chose the weights $\mx{S} = \mathrm{diag}(5,5,10,10,10,10)$ and
$\mx{T} = \mathrm{diag}(1000,\cdots,1000)$ for the \gls{qp} \eqref{eq:qp_wbc}.
The parameters and weights used by the \gls{nmpc} are reported in Table 
\ref{tab:nmpc_parameters} and \ref{tab:nmpc_weights}, respectively. 
In all of our simulations and experiments, we do not set any weights on the 
\gls{com} position ($\vect{p}_\mathrm{c}$), roll ($\phi$) and pitch ($\theta$) tasks because we wanted 
the \gls{nmpc} to sort out these \gls{com} trajectories autonomously. 
%
%
%
%
%
%
%
\begin{table}[h!]
	\caption{ \gls{nmpc} parameters}
	\begin{center}
		\begin{tabular}{@{} l l l l @{}}
			\toprule[0.4mm]
			\textbf{Parameter} & {\textbf{Symbol}} & \textbf{Value}  & \textbf{Unit} \\ 
			\midrule
			Number of state & $n_x$&  12  & -   \\
			Number of control inputs & $n_u$&  12  & -    \\
			Number of model parameters & $n_a$&  16  & -    \\
			Prediction horizon & $T$&  2  & $\mathrm{s}$    \\
			Sampling time &$T_\mathrm{s}$&  0.04  & $\mathrm{s}$    \\
			Number of control intervals & $N$&  50  & -    \\
			Friction coefficient & $\mu$	&  0.7     &  - \\	
			\gls{grfs} lower bound & $\underline{\vect{f}}_\mathrm{z}$&  0  & $\mathrm{N}$    \\
			\gls{grfs} upper bound &  $\overline{\vect{f}}_\mathrm{z}$&  500  & $\mathrm{N}$    \\
			Hip-to-foot distance reference & ${}_\mathcal{C}\vect{p}_{\mathrm{hf},i}^{\mathrm{ref}}$&  ($0.0, 0.0, -0.55$)  & $\mathrm{m}$ \\
			Regularization parameter & $\rho$& $3\times 10^{-5}$  & -  \\ 
			\bottomrule[0.4mm]
		\end{tabular}
		\label{tab:nmpc_parameters}
	\end{center}
\end{table}
\begin{table}[h!]
	\caption{ Weights used in the \gls{nmpc}}
	\begin{center}
		\begin{tabular}{@{} l l l @{}}
			\toprule[0.4mm]
			\textbf{Cost} & \textbf{Weight} & \textbf{Value} \\ 
			\midrule	
			\multirow{4}{4em}{State} 			
			&$\mx{Q}_{\vect{p}_\mathrm{c}}$  & $\mathrm{diag}$($0$, $0$, $0$)   \\
			&$\mx{Q}_{\vect{v}}$	 & $\mathrm{diag}$($100$, $100$, $100$) \\
			&$\mx{Q}_{\bm{\Phi}}$	  & $\mathrm{diag}$($0$, $0$, $100$) \\
			&$\mx{Q}_{\bm{\omega}}$ 	  & $\mathrm{diag}$($100$, $100$, $1000$)  \\
			\midrule
			\multirow{3}{4em}{Force}
			&$\mx{R}_\mathrm{x}$ 	  & $1\times 10^{-3}$    \\
			&$\mx{R}_\mathrm{y}$ 	  & $1\times 10^{-3}$    \\
			&$\mx{R}_\mathrm{z}$	  & $8\times 10^{-4}$ \\
			\midrule
			\multirow{3}{4em}{Mobility}
			&$\mx{M}_\mathrm{x}$ 	  & $1\times 10^{-4}$    \\
			&$\mx{M}_\mathrm{y}$ 	  & $2\times 10^{-3}$    \\
			&$\mx{M}_\mathrm{z}$	  & $1000$ \\
			\midrule
			\multirow{3}{4em}{Force robustness}
			&$\mx{P}_\mathrm{x}$ 	 &  $100$\\
			&$\mx{P}_\mathrm{y}$ 	 &  $100$ \\
			&$\mx{P}_\mathrm{z}$	 &  $1$  \\
			\bottomrule[0.4mm]
		\end{tabular}
		\label{tab:nmpc_weights}
	\end{center}
\end{table}
\subsubsection{Discretization}
For the discretization of the dynamic constraints,  
we mainly investigated two integration schemes, i.e, a single step of the explicit Euler of order $1$ and implicit midpoint method of order $2$ due to their low computational complexity that 
favors our real-time implementation needs. 
We chose the 
implicit midpoint method of order $2$ because of its stability and accuracy properties.
The sampling time $T_\mathrm{s}=40 \,\mathrm{ms}$ was chosen and it was sufficient to conduct 
\gls{nmpc} computation \textit{online} along with the other necessary computations for the re-planning.
\subsubsection{\gls{nmpc} software}
We use the $\texttt{acados}$ software package \cite{Verschueren2019} to implement 
the \gls{rti} scheme described in Section \ref{sec:rti_scheme}. Since $\texttt{acados}$ comes 
with a Python interface allowing rapid prototyping, we first tuned the algorithm in
simulation and then used the generated $\texttt{C}$-code to perform real experiments.
%
%
We employ the \gls{qp} solver High-Performance Interior Point Method (HPIPM) \cite{Frison2020}, 
which exploits the sparsity structure of the MPC QP sub-problem \eqref{eq:qp_subproblem}, 
and supports inequality constraints. 

The computation time required by the \gls{nmpc} was in the range of $5$-$7 \, \mathrm{ms}$ 
with the prediction horizon of $2 \, \mathrm{s}$ and control intervals $N$ equal to $50$ on the 
on-board computer (a Quad Core Intel Pentium PC104 @ 1GHz) of \gls{hyq} for all the experiments. 
This computation time corresponds to the feedback phase of the \gls{rti} scheme where 
the \gls{qp} \eqref{eq:qp_subproblem} is solved after receiving the current state of the robot.
The preparation phase of the \gls{rti} takes about $2$-$3 \, \mathrm{ms}$ which 
is a fraction of the sampling time we chose. 
Refer to Section \ref{sec:rti_scheme} for more details on these phases of the \gls{rti} scheme. 
Even though the computation time of \gls{nmpc} is mostly consistent, we observed some outliers.
Hence we opted for a conservative approach to run the \gls{nmpc} at $25\,\mathrm{Hz}$ 
to guarantee that the computation time stays always less than $40 \, \mathrm{ms}$. Besides the 
computation time of the \gls{nmpc}, we also account for the time required by other blocks
such as reference generator so that the total computation time does not exceed $40 \, \mathrm{ms}$. 

%
%
%
%
\subsubsection{Integration with the locomotion framework}
The \gls{nmpc} is integrated in a ROS node that publishes  $\vect{x}^\mathrm{p}$ and $\vect{u}^\mathrm{p}$ 
at a frequency of 25 $\mathrm{Hz}$. The on-board computer along with our locomotion framework (\gls{wbc} 
Interface, \gls{wbc}, etc., illustrated in Fig. \ref{fig:blockDiagram})
runs a real-time node that subscribes to the topic of the \gls{nmpc} ROS node. 
ROS is not a real-time operating system, so it can introduce quite a significant and unpredictable communication delay if the  \gls{nmpc}
is run on an external (e.g. more powerful) computer. 
These delays are difficult to compensate for and they can cause a loss of 
synchronization between the \gls{nmpc} ROS node 
and the \gls{wbc} interface.
Therefore, we decided to launch the \gls{nmpc} node natively on the 
on-board computer to avoid communication delays between two different computers. 
Even though we chose not to use a more powerful dedicated off-board computer for the \gls{nmpc} ROS node, 
we obtained a better performance in the overall implementation by avoiding the communication delays of ROS.
\subsection{Simulations}
We show our \gls{nmpc} planner in action on challenging terrain starting 
with simulations. These simulations are pallet crossing, walking over 
unstructured rough terrain and walk into a V-shaped Chimney.
\subsubsection{Pallet crossing}
\label{sec:pallet_cross}
\begin{figure}[!h]
	\centering
	\includegraphics[width=\columnwidth]{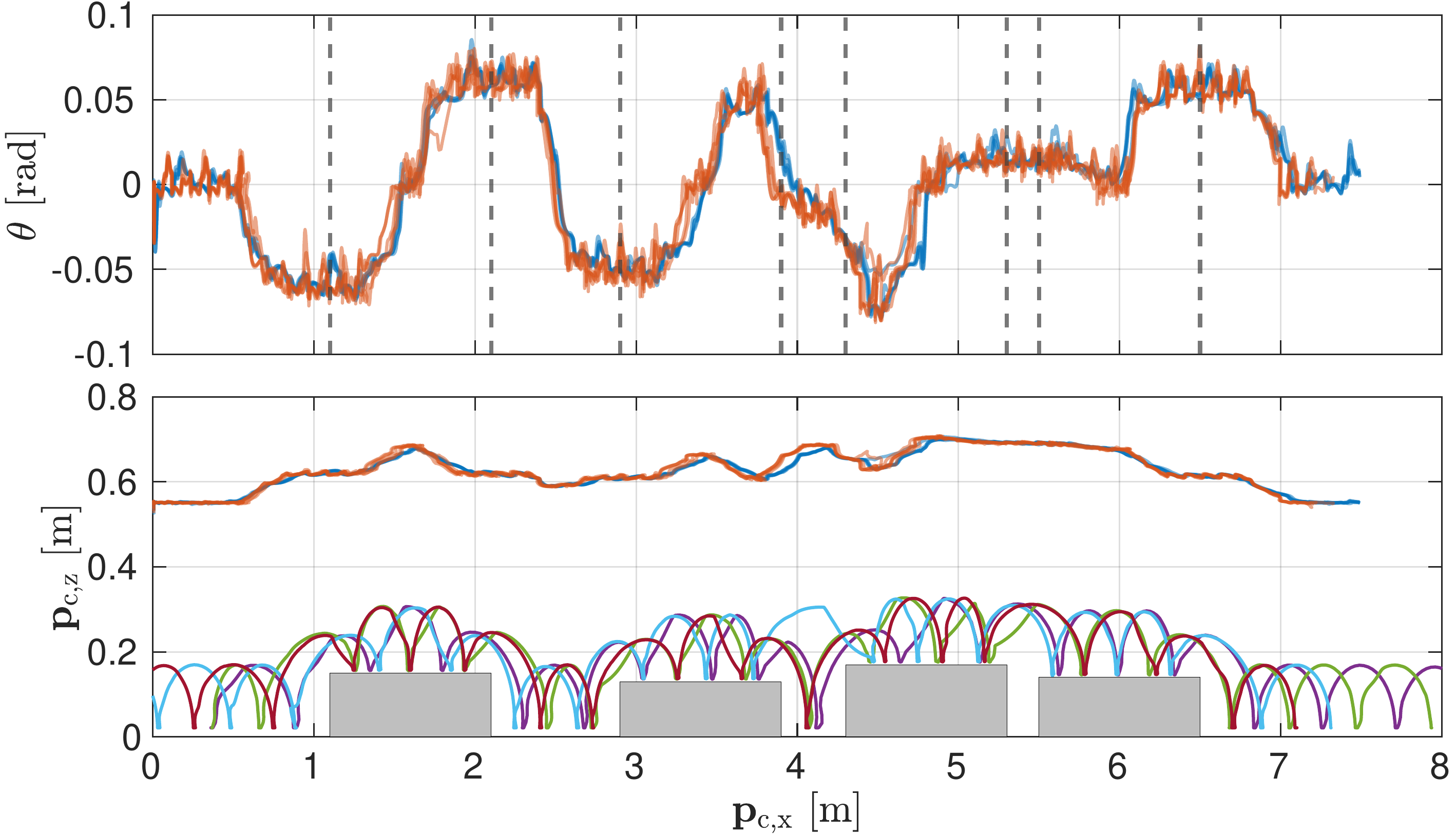}
	\caption{ Simulation of pallet crossing scenario for five different trials commanding the robot to cross at 0.1~$\mathrm{m/s}$~(blue) and 0.05~$\mathrm{m/s}$~(red). The top graph shows the 
		actual velocities (solid lines) and their respective references (dashed lines). 
		The graph in the middle shows the pitch angle and the dashed vertical lines indicate the edges of the pallets for that specific 
		location in the plot. The bottom graph shows the \gls{com} $Z$ position
		for all the trials and the feet trajectories for one of the trials performed at 0.1 $\mathrm{m/s}$. The color of swings are related to the different legs.}
	\label{fig:multi_pallet}
\end{figure}

In this simulation \gls{hyq} traverses pallets of 
different heights, placed at varying distances form each other. This simulation highlights 
the importance of including mobility in the \gls{nmpc} formulation \eqref{eq:mobility_cost}. 
In particular, the simulation scenario includes a set of pallets, each one of $1 \, \mathrm{m}$ length, 
with variable heights between $0.13$ and  $0.17\, \mathrm{m}$ 
and placed at unequal gap lengths ranging from $0.2$ and  $0.7\,\mathrm{m}$. 
We performed multiple trials commanding the robot to move forward at 
different velocities i.e.,  $0.05\,\mathrm{m/s}$ and  $0.1\, \mathrm{m/s}$ to show the repeatability of our approach. 
To avoid stepping on undesired locations such as pallet edges and to prevent foot or shin collisions,
the nominal footholds were  adjusted by using the~\gls{vfa} (see Section \ref{sec:robocentric_stepping}). 
Fig.~\ref{fig:multi_pallet} shows the results of five different trials for 
each of the commanded velocities. The middle plot shows the pitch angle $\theta$ of the robot as it 
traverses the scenario.  Since the 
robot is only commanded to move along its $X$ direction with a constant forward 
velocity and the foot locations are provided as known quantities 
to the \gls{nmpc}, the adjustment in pitch is the result of minimizing the deviation
from the hip-to-foot distance configuration corresponding to high mobility for 
all four legs. 
Without this feature, 
the robot would maintain a constant horizontal orientation eventually reaching low mobility
in some legs as shown in the attached video\footnote{https://youtu.be/r0-KIiw0eWM} for a single pallet simulation. 

We also showed in the accompanying video the simulation of a walk on 
randomly generated rough terrain (using terrain generation tool by \cite{Bulat2020}) with the forward velocity of $0.3 \, \mathrm{m/s}$ 
further stressing the advantages of mobility cost mentioned earlier. 
\subsubsection{Walk into a V-shaped chimney}\label{sim:vshaped_wall}
In this simulation we show \gls{hyq} walking at $0.03\, \mathrm{m/s}$ commanded velocity in the $X$ direction into a
V-shaped chimney with friction coefficient $\mu= 0.7$ and walls inclined at $35^\circ$ to the ground. 
This simulation exploits the cone constraints and force robustness cost defined inside the
\gls{nmpc} formulation that is vital for the success of this task.
The robot receives online an update of the map of the environment through an on-board camera to get the 
information about the normals at the location of the contact. These normals are used to formulate the force
robustness cost \eqref{eq:force_robustness} in the contact frame $\mathcal{K}$.
With this cost, the NMPC provides optimal \gls{grfs} to stay close to the normals 
of the friction cones at the contacts. 

\begin{figure}[!h]
	\centering
	\includegraphics[width=\columnwidth]{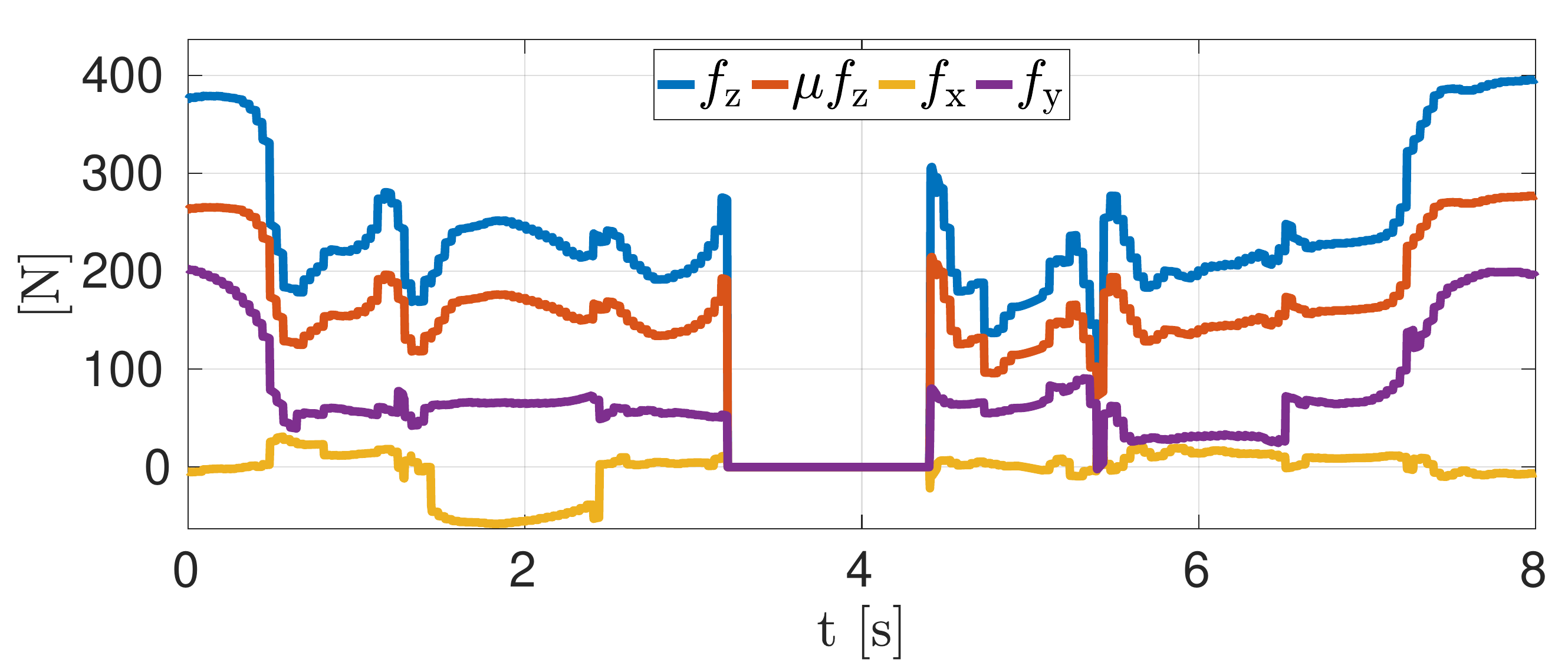}
	\caption{ Walk into a V-shaped chimney simulation: \gls{grfs} of \gls{lf} 
		leg for a single gait cycle with cone constraints 
		and regularization cost. Both the longitudinal $f_\mathrm{x}$ and 
		lateral $f_\mathrm{y}$ lie conservatively within the bound $\mu f_\mathrm{z}$ imposed by cone constraints.}
	\label{fig:friction_cone_forces}
\end{figure}

\begin{figure}[!h]
	\centering
	\includegraphics[width=\columnwidth]{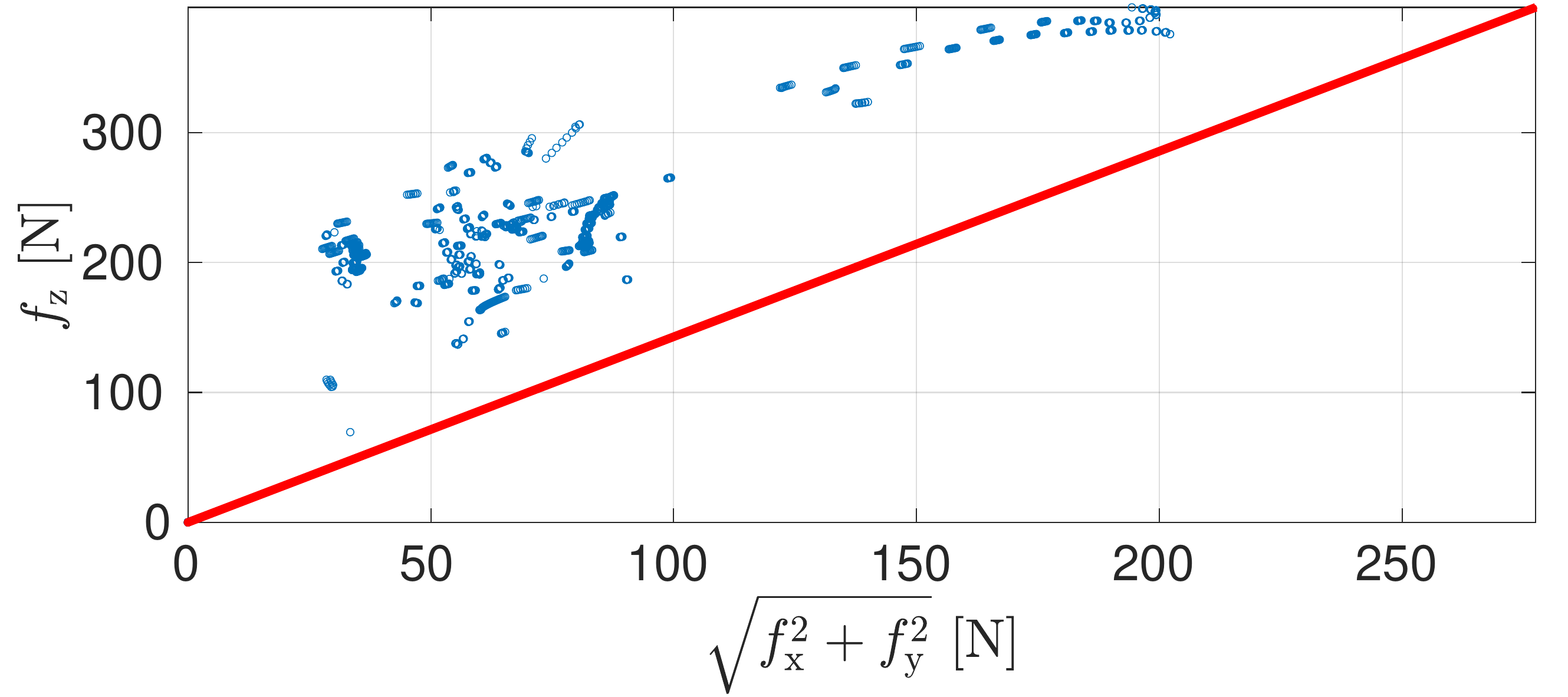}
	\caption{ Walk into a V-shaped chimney simulation: Normal force $f_z$ versus tangential force $\sqrt{f_\mathrm{x}^2 + f_\mathrm{y}^2}$ of the \gls{lf} leg for a single gait cycle
		expressed in the contact frame. The red line is the cone bound $\mu f_\mathrm{z}$.}
	\label{fig:friction_cone_violation}
\end{figure}

As shown in the accompanying video, without the cone constraints the robot  
slips while climbing the chimney and ultimately falls. When the force robustness cost is enabled, the 
forces are regularized to stay in the middle of the cones, thanks to the robustness 
feature described in Section \ref{subsec:force_robustness}. In this case the robot walks successfully 
into the chimney.
In Fig. \ref{fig:friction_cone_forces}, it can be seen that the longitudinal and lateral
components of the GRF at the \gls{lf} foot stay within the bound $\mu f_\mathrm{z}$ (in red) imposed 
by the cone constraints. Moreover, Fig. \ref{fig:friction_cone_violation} 
plots the normal versus the tangential force of the GRF 
together with the cone bound $\mu f_\mathrm{z}$ (red line). The picture shows that 
the GRF stays well within the bound without any violation. 
Therefore, including the force regularization term 
enables the \gls{nmpc} to account for the estimation error in the orientation of
contact normals and increase robustness to the external disturbances. 
%
%

\begin{figure}
	\centering
	\includegraphics[width=\columnwidth]{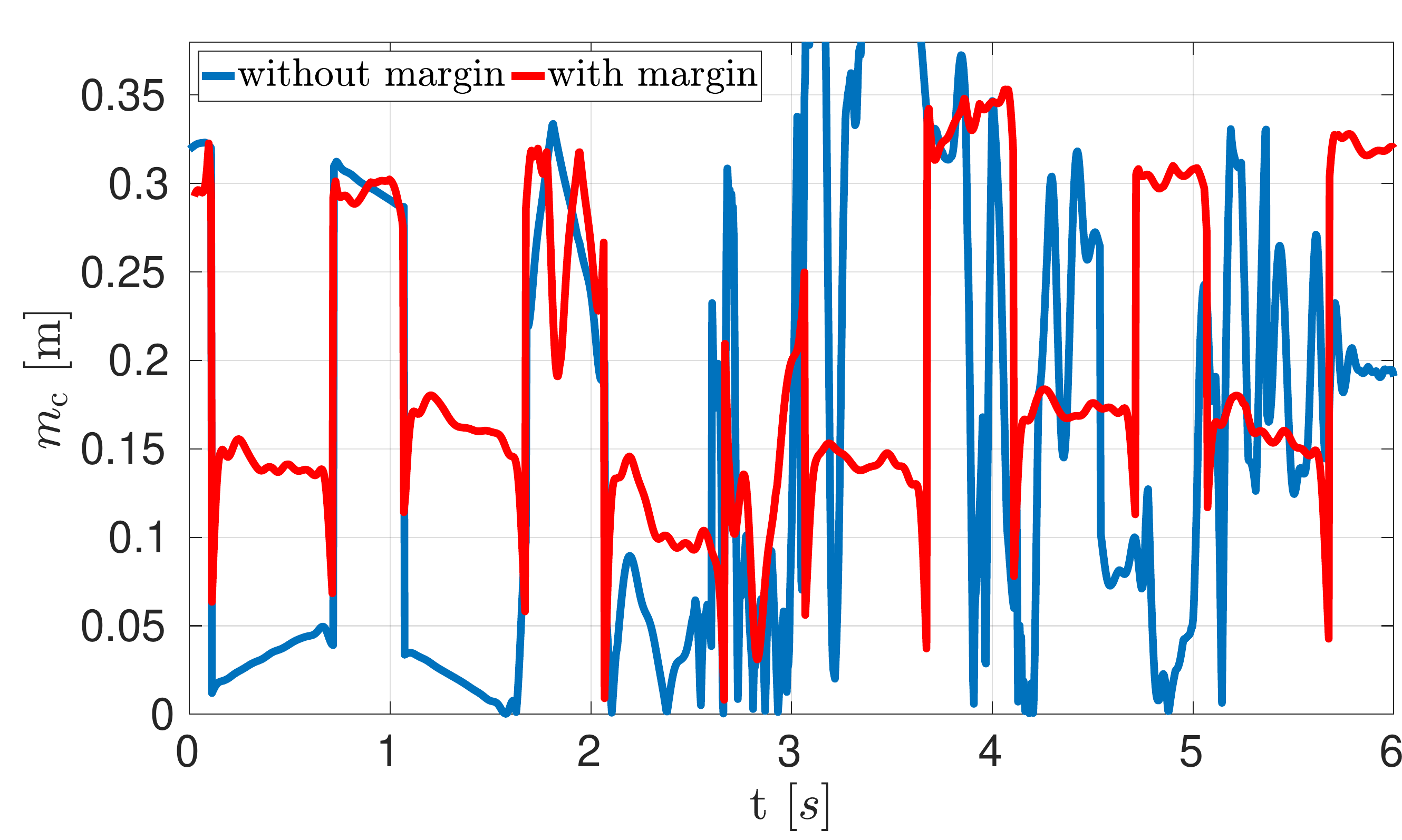}
	\caption{ Plot of the \gls{zmp} margin $m_\mathrm{c}$ used to measure locomotion stability. 
		The robot is pushed immediately after $2\,\mathrm{s}$ with lateral force of $200\, \mathrm{N}$ while
		 walking on a flat terrain at  $0.1\,\mathrm{m/s}$ \gls{com} $X$ velocity. 
		 The discontinuities are due to the switching between 3/4 stance legs in a crawl gait.}
	\label{fig:stability_margin}
\end{figure}

Apart from the simulation mentioned above, we also have added in the attached video, 
the simulations regarding the \gls{zmp} margin (see Section \ref{subsec:zmp_margin}) 
and the importance of the re-planning at a higher rate.
For the \gls{zmp} margin simulation, the robot is pushed with $200\, \mathrm{N}$ of 
lateral force for $1 \, \mathrm{s}$ both in case of sufficient (higher weight on \gls{grfs} $Z$) 
and no (lower weight on \gls{grfs} $Z$) \gls{zmp} margin. The \gls{zmp} margin 
plots for this simulation can be seen in Fig. \ref{fig:stability_margin}
where the \gls{zmp} margin is improved in case of 
the red line compared to the blue one because the \gls{grfs} 
$Z$ components are penalized relatively more ($100$ times)  for the red line.
Because of the improved margin the robot walks stably, whereas
it falls while walking when there is no margin. 

In the second simulation, the robot is commanded with a constant \gls{com} X velocity 
and heading velocity simultaneously. In case of re-planning at lower rate of $0.8\, \mathrm{Hz}$,
the robot becomes unstable and falls due to increase in the model uncertainties and tracking errors. 
We would like to stress that when the re-planning is done at a lower
frequency than $25\,\mathrm{Hz}$, the robot is in open-loop for the time interval 
between two consecutive re-planning instances, hence, it is no more \gls{nmpc}  
but an online open-loop trajectory optimization. On the other hand, at a higher 
re-planning frequency of $25\,\mathrm{Hz}$ the robot walks successfully because
the \gls{nmpc} compensates for the model uncertainties and tracking errors.

\subsection{Experiments}
We performed three different experiments to demonstrate the 
real-time implementation of our \gls{nmpc} running on the 
on-board computer of the robot as follows.
\subsubsection{Omni-directional walk} \label{subsec:omnidirectional}
\begin{figure}[!h]
	\centering
	\includegraphics[width=\columnwidth]{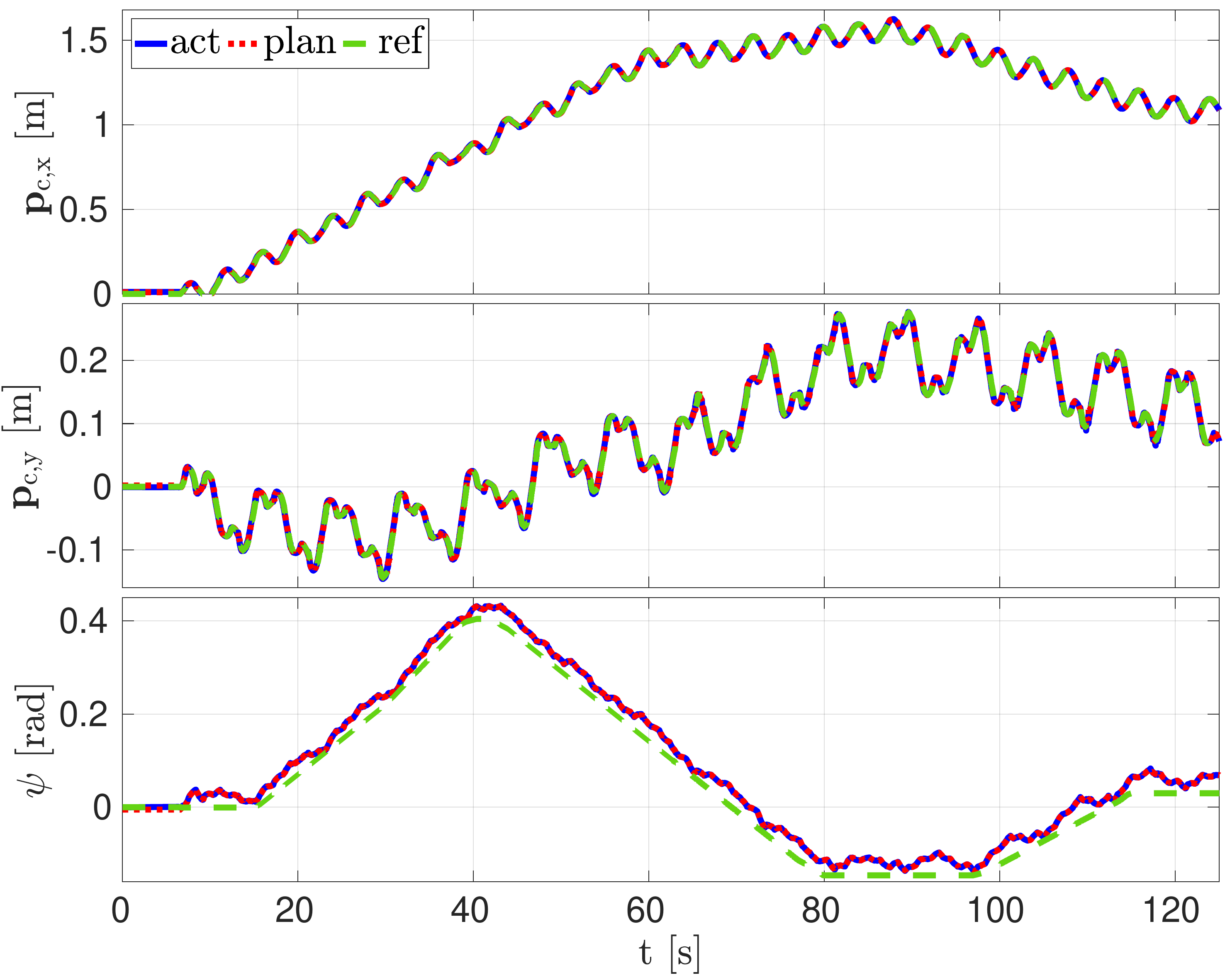}
	\caption{ \gls{com} X-Y position and  
		yaw $\psi$ in omni-directional walk experiment. The blue, dotted red and dashed 
		green line represent the actual, planned and reference values, respectively.}
	\label{fig:omni_walk_comPos}
\end{figure}

\begin{figure}[!h]
	\centering
	\includegraphics[width=\columnwidth]{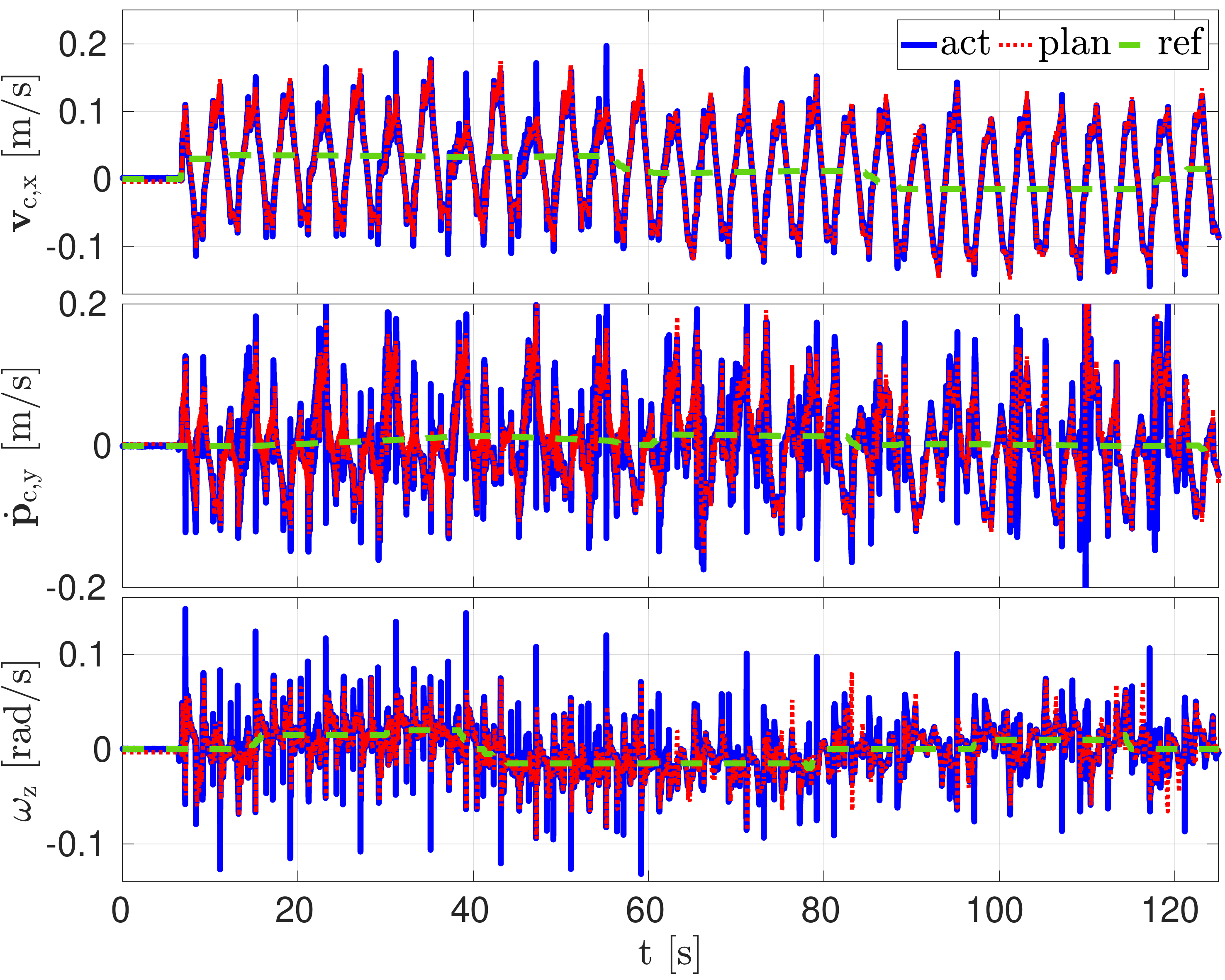}
	\caption{ The longitudinal $\dot{\vect{p}}_\mathrm{c,x}$, lateral $\dot{\vect{p}}_\mathrm{c,y}$  and angular $\bm{\omega}_\mathrm{z}$ velocity  of the robot in omni-directional walk
		experiment.  The blue, dotted red and the green line represent the actual,
		planned and reference values, respectively.}
	\label{fig:omni_walk_velocity}
\end{figure}

\begin{figure}[!h]
	\centering
	\includegraphics[width=\columnwidth]{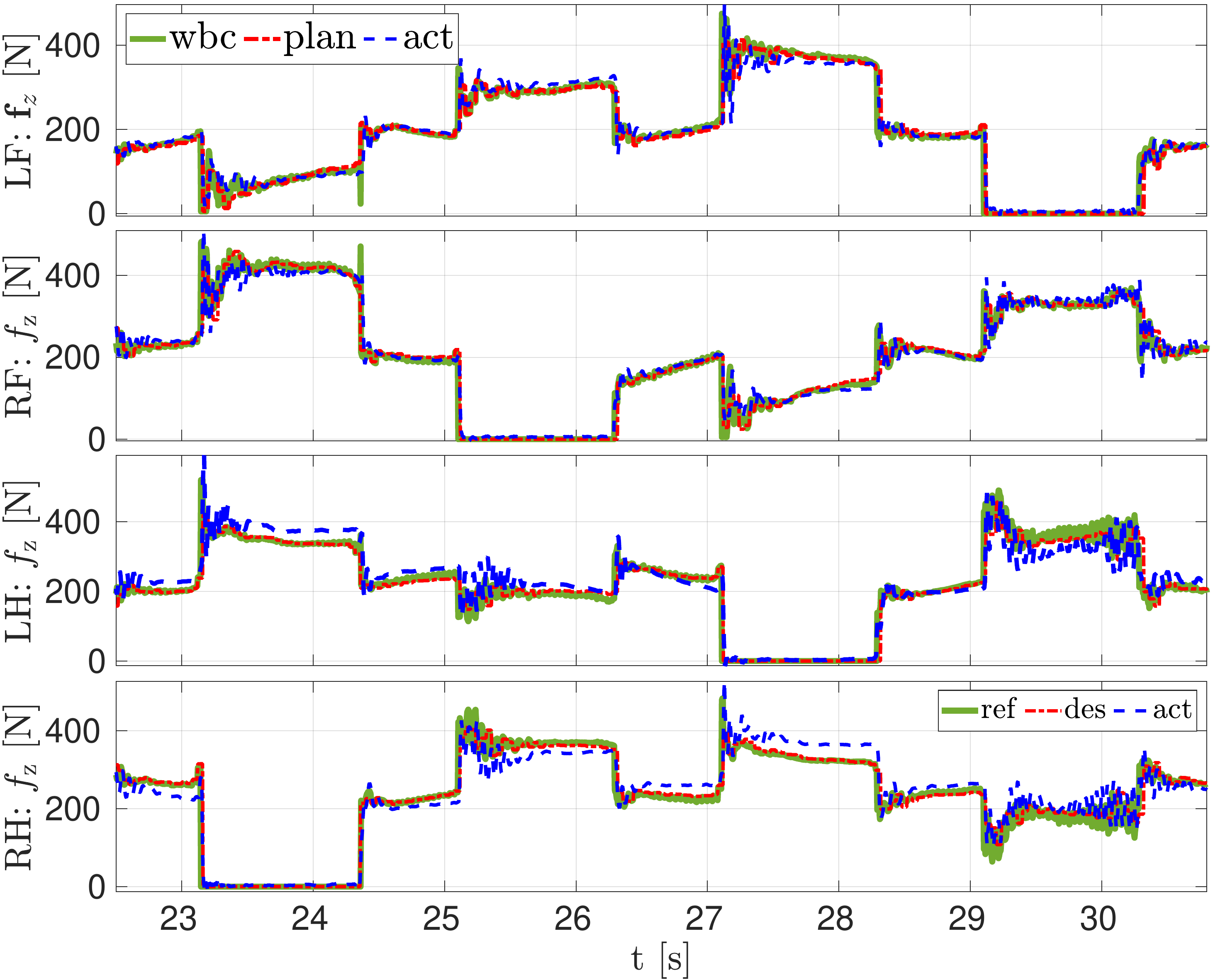}
	\caption{ \gls{grfs} of one gait cycle in the omni-directional walk experiment
		(We show only one cycle for better visibility of the data). The green,
		dotted red and dashed blue line represent the output from \gls{wbc}
		i.e., $f_{\mathrm{z},i}^\mathrm{d}$, planned and reference values, respectively.}
	\label{fig:omni_walk_forces}
\end{figure}

With this experiment, we show the omni-directional walk 
performed by HyQ with the \gls{nmpc} on a flat terrain. This experiment
validates that the \gls{nmpc} computes feasible trajectories after receiving
different velocity commands from the user while walking.
In this experiment, the robot was commanded with a longitudinal velocity 
${}_\mathcal{H} \vect{v}_\mathrm{c,x}^{\mathrm{usr}}$ by the user to walk forward/backward 
and then a lateral velocity ${}_\mathcal{H} \vect{v}_\mathrm{c,y}^{\mathrm{usr}}$. Finally, a heading velocity $\omega^{\mathrm{usr}}_\mathrm{z}$ was commanded  
to turn in the left/right direction.
Fig.~\ref{fig:omni_walk_comPos}  
shows the \gls{com} $X$-$Y$ position
and yaw angle of the robot base and it can be noticed that the actual values 
track very closely the planned trajectories provided by the \gls{nmpc}. 
Fig.~\ref{fig:omni_walk_velocity} 
depicts the deviation of the actual velocities from the reference values
while following the planned trajectories form \gls{nmpc}.
It can be seen in Fig.~\ref{fig:omni_walk_forces} that
the \gls{grfs} generated by the \gls{wbc} are compliant with the planned values 
$\vect{u}^\mathrm{p}$ and again the actual values of \gls{grfs} 
track closely the planned values. From these plots, it can be observed that the 
continuous re-planning with \gls{nmpc} plays an important role to achieve
good tracking of the planned trajectory.

%
%
\subsubsection{Traversing a static pallet}\label{sub:traversing_pallet}

\begin{figure}[!h]
	\centering
	\includegraphics[width=\columnwidth]{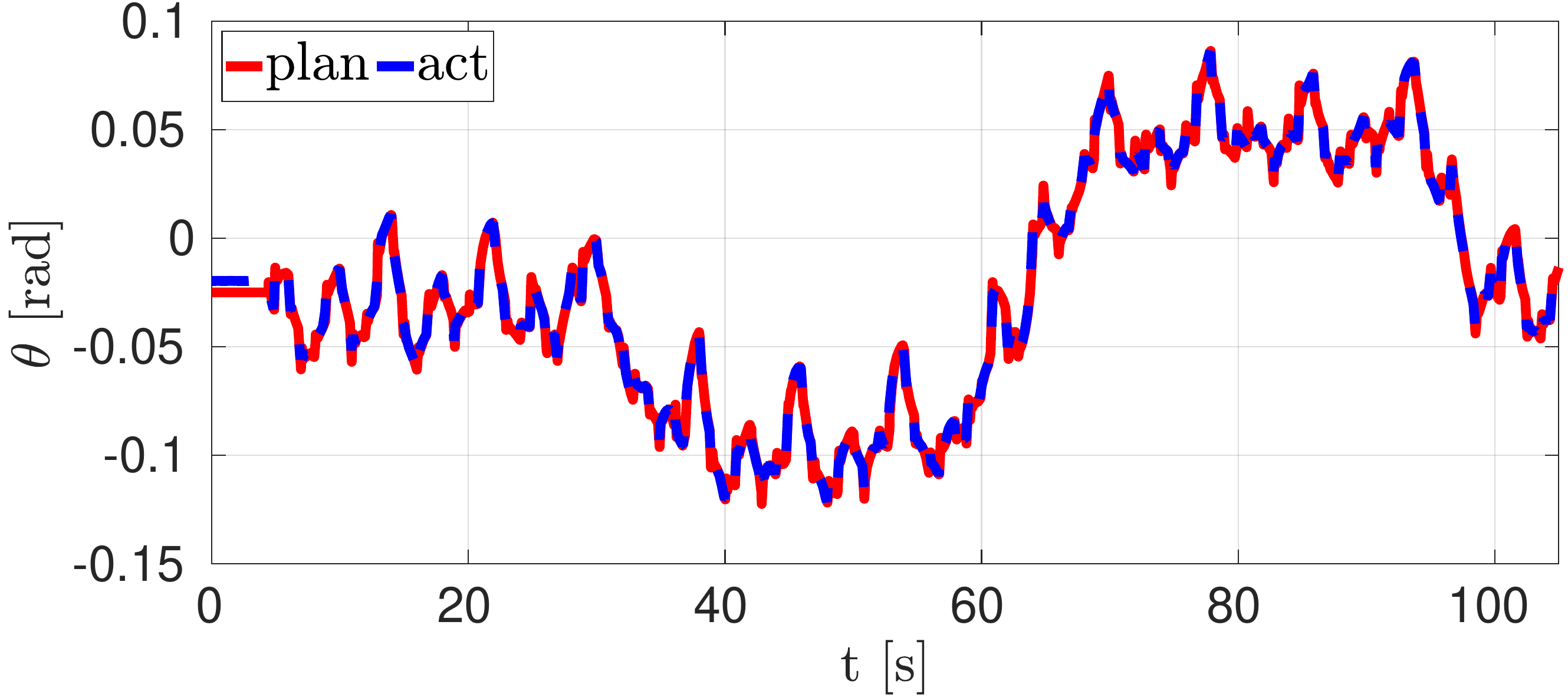}
	\caption{ Planned (red) and Actual (dashed blue) pitch of the robot
		base while traversing a static pallet in the experiment at  
		a commanded \gls{com} $X$ velocity of $0.03 \, \mathrm{m/s}$.}
	\label{fig:staticPallet_pitch}
\end{figure}

The purpose of this experiment was to demonstrate that the mobility cost \eqref{eq:mobility_cost}
incorporated in the \gls{nmpc} formulation provides the necessary body pitch  for the robot 
to traverse over a static pallet while maintaining good leg mobility. The pallet used in this experiment was $0.13\,\mathrm{m}$ in height and $0.8\,\mathrm{m}$ in length.
Fig.~\ref{fig:staticPallet_pitch} shows that the robot 
pitches up while climbing up the pallet and pitches down consequently while climbing 
down from the pallet. As shown in the attached video in simulation, when the mobility cost is deactivated, 
the \gls{nmpc} maintains the horizontal base orientation. This causes a reduced hip-to-foot distance
while stepping up/down on the pallet ultimately resulting in low leg mobility. When mobility 
cost is activated, it directs the \gls{nmpc} solution to achieve the necessary pitch that allows
to maintain the hip-to-foot distance at the reference value and hence the leg mobility is improved. 
Moreover, the \gls{vfa} provides the
corrected foot position (i.e., to avoid shin or feet collisions with the edges of the pallet) to the \gls{nmpc} 
and this further enhances the overall locomotion.
%
%
\subsubsection{Traversing a repositioned pallet } 
\begin{figure}[!h]
	\centering
	\includegraphics[width=\columnwidth]{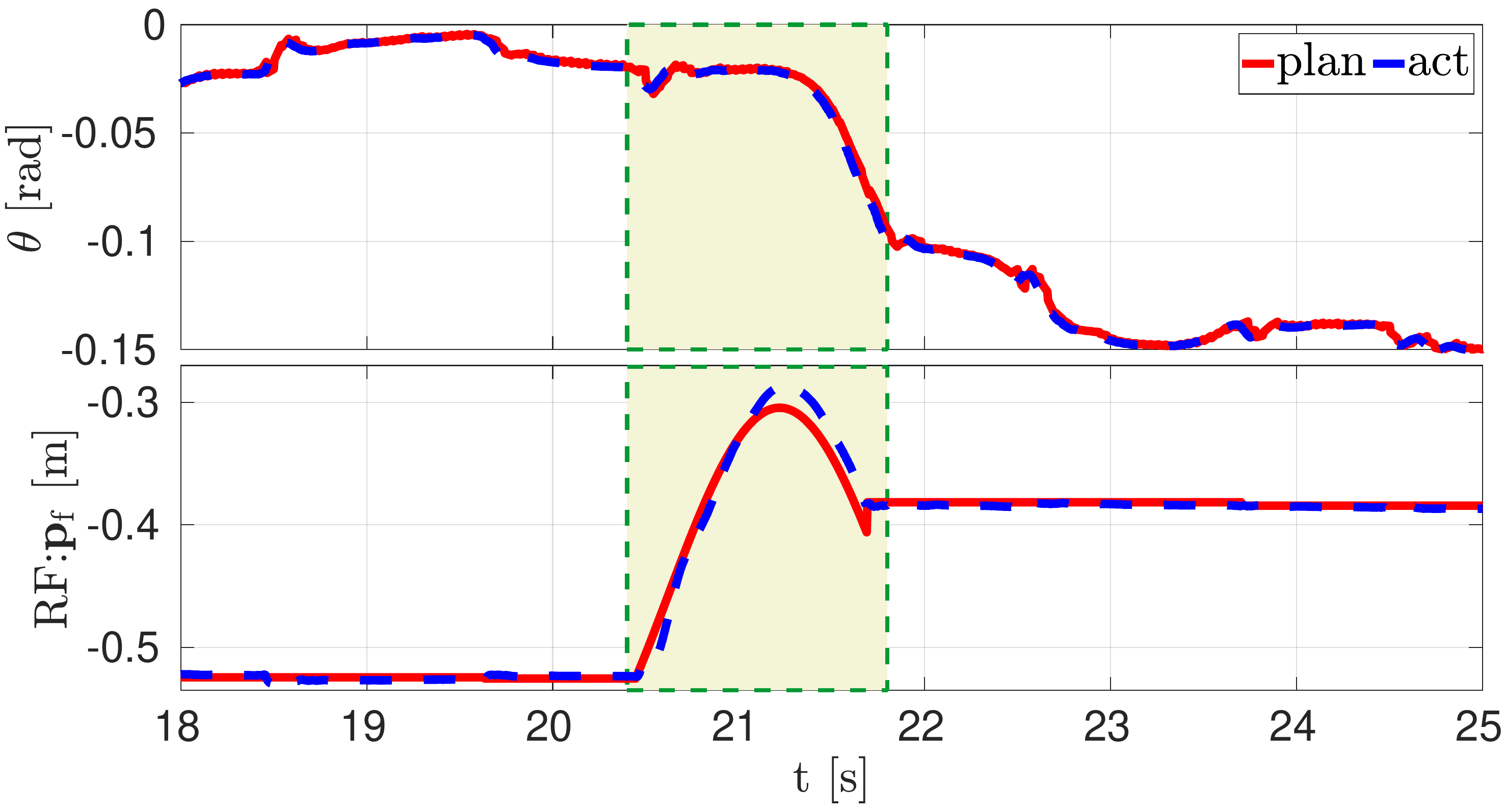}
	\caption{ Robot base pitch achieved during the swing of \gls{rf} 
		leg while traversing a repositioned pallet in the experiment a commanded \gls{com} $X$ velocity of $0.05 \, \mathrm{m/s}$.
		The red and dashed blue line are planned and actual values.}
	\label{fig:pitch_while_swing}
\end{figure}

In this experiment we tested our \gls{nmpc} to plan the robot motion in real-time
by adapting the changes in the environment with the help of \gls{vfa}.
As it can be seen from the attached video, when the pallet ($0.13\,\mathrm{m}$ in height and $0.8\,\mathrm{m}$ in length) is pushed in 
front of \gls{hyq} while walking, the heightmap detects the pallet 
and the \gls{vfa} provides updated foot locations to the \gls{nmpc}.
The \gls{nmpc} after receiving these updated foot locations
delivers a solution by pitching up the robot base in order to adapt to 
the change in the environment while maintaining the mobility.
Even though the mobility cost is
defined for the stance legs, it is interesting to notice that the \gls{nmpc} decides to adjust the base pitch while swinging the \gls{rf} leg onto the pallet (see Fig. \ref{fig:pitch_while_swing})
by forecasting the  change in hip-to-foot distance at the touchdown. 
This experiment highlights the advantage of the predictive control
over the traditional control apporaches for its ability to incorporate the 
knowledge of the future states. It also validates
the effectiveness of our mobility cost in the \gls{nmpc} coupled with 
the \gls{vfa} to adapt to the changes in the locomotion environment.
%
%
%
%

\section{Conclusion}\label{sec:conclusion}
In this work, we have demonstrated in experiments a real-time \gls{nmpc} which leverages optimization of leg mobility to achieve terrain adaptation. 
The contact sequence parameters embedded inside the \gls{srbd} model allows
us to encode the complementarity constraints directly, and without a need to enforce these constraints separately in the \gls{nmpc}.
We exploited the \gls{rti} scheme for our \gls{nmpc} that enable us to close the loop 
at  $25\,\mathrm{Hz}$ on the \gls{nmpc} with a
prediction horizon of $2 \,\mathrm{s}$. Closing the loop on \gls{nmpc} at  $25\,\mathrm{Hz}$ 
allows us to compensate for the state drifts due to model 
uncertainties and tracking errors, and also adapt to the changes in 
the environment while following user velocity commands both in the simulations
and experiments. 


%
%

In our \gls{nmpc}, the  mobility cost penalizes the hip-to-foot distance from a 
reference value corresponding to a high mobility factor and hence it
directs the \gls{nmpc} to compute essential robot orientation to maintain 
a high mobility while respecting the kinematic limits. 
This is evident from the pallet experiments where we also included the \gls{vfa} 
to correct undesired foot positions defined by the heuristics and avoid possible foot and shin collision. 
Accounting for the \gls{zmp} margin in our \gls{nmpc} improved the 
locomotion stability of the robot in all of our experiments and 
simulation by keeping a sufficiently large \gls{zmp} margin from support polygon boundaries. 
Incorporating a force robustness term in the \gls{nmpc}
ensures that the \gls{grfs} stay close to the contact normals and hence, it enables the robot to cope with the estimation error of the orientation of the contact normals. 
%
%


With our \gls{nmpc}, we have performed successful dynamic locomotion in simulation as 
well as in the experiments on different rough terrains.
In our future work we would like 
to extend our \gls{nmpc} to optimize the step timing and foot locations.  Additionally, 
the reference generator does not provide references by rejecting the external disturbances
acting on the robot state, hence the robot complies transparently 
with these disturbances. Therefore, in the future we plan to  empower the reference 
generator to reject disturbances to bring back the robot from a perturbed state to a state coherent with the user commands.
\appendices

\section{Angular Velocity} \label{adx:eulerRate}
We employ the  $Z$-$Y$-$X$ convention \cite{Diebel2006} for the Euler angles sequence 
$\bm{\Phi}=(\phi,\, \theta,\, \psi)^\top$ to represent the orientation of the 
robot base  where, $\phi$, $\theta$ and $\psi$  are the roll, pitch and yaw, respectively. 
The angular velocity in inertial and \gls{com} frame is related 
to the Euler angle rates with the following relations:
\begin{align}
\bm{\omega} &=  \mx{E}(\bm{\Phi})\,\dot{\bm{\Phi}}\\
{}_\mathcal{C}\bm{\omega} &=  \mx{E}'(\bm{\Phi})\,\dot{\bm{\Phi}}
\end{align}
$\mx{E}(\bm{\Phi})$ and $\mx{E}'(\bm{\Phi})$ are the \textit{Euler angle rates matrix}
and \textit{conjugate Euler angle rates matrix}  respectively given by,

\begin{equation}
\mx{E}(\bm{\Phi})=
\mat{ \cos(\theta)\cos(\psi) &  -\sin(\psi) &  0 \\
	\cos(\theta)\sin(\psi) &  \cos(\psi) &  0 \\
	-\sin(\theta)			 &  0		   &  1 }
\end{equation}
\begin{equation}
\mx{E}'(\bm{\Phi})=
\mat{ 1 &  0 		  &  -\sin(\theta) \\
	0 &  \cos(\phi) &  \cos(\theta)\sin(\phi) \\
	0 &  -\sin(\phi) &  \cos(\theta)\cos(\phi) }
\end{equation}
Remark: $\mx{E}$ depends on pitch and yaw, whereas $\mx{E}'$ on roll and pitch.
Thus, the Euler angle rates $\dot{\bm{\Phi}}$ is
\begin{align}
\dot{\bm{\Phi}}&=\mx{E}^{-1}(\bm{\Phi})\,\bm{\omega}\\
\dot{\bm{\Phi}}&=\mx{E}'^{-1}(\bm{\Phi})\,{}_\mathcal{C}\bm{\omega}
\end{align}
%
%



\section*{Acknowledgment}
We would like to thank Jonathan Frey and Andrea Zanelli of IMTEK, University of 
Freiburg for their support with $\texttt{acados}$ software. We also extend our
gratitude to all the DLS lab members for their help in  the experiments. 

\small
\bibliographystyle{IEEEtran}
\bibliography{references/references}
\begin{IEEEbiography}[{\includegraphics[width=1in,height=1.25in,clip,keepaspectratio]{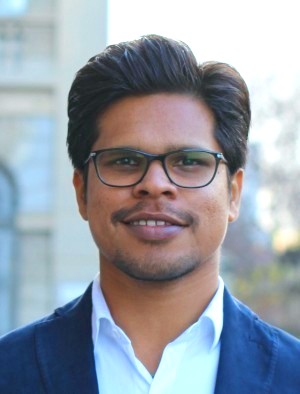}}]{Niraj Rathod}
 received the Bachelor's degree in Instrumentation and Control from College of Engineering Pune, India, in 2011. 
 Immediately after bachelor's he worked as a System Engineer in EEEC, Pune, India, until September 2014. 
 Thereafter, he studied the Master's in Automation and Control Engineering from Politecnico di Milano, Italy, and graduated in 2017. For his Master's thesis he collaborated with RSE S.p.A., Milan, Italy. 
 He is currently a PhD student at IMT School for Advanced Studies Lucca in Computer Science and System Engineering. 
 During his PhD, he was an Affiliated Researcher at Dynamic Legged System (DLS), IIT, Genova, Italy. 
 His research interests include MPC, Optimal Control, Control of Legged Robots and Hybrid Systems.
\end{IEEEbiography}
\vspace{-0.9cm}
\begin{IEEEbiography}[{\includegraphics[width=1in,height=1.25in,clip,keepaspectratio]{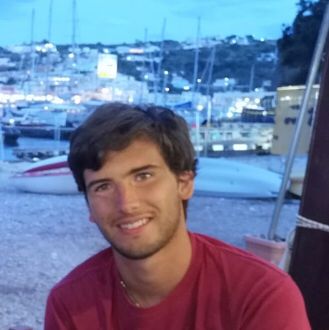}}]{Angelo Bratta}
	received his B.Sc. degree in control Engineering from the  Politecnico di Bari, in 2017,
	and the M.Sc. degree in Mechatronics Engineering from the Politecnico di Torino, in 2019.
	In March 2019, he joined the Dynamic Legged	Systems (DLS) lab at Istituto Italiano di Tecnologia
	(IIT) for his master thesis.
	Since November 2019 he is a Ph.D. student of the DLS. His research interests include robotics, controls,
	and online trajectory optimization.
\end{IEEEbiography}
\vspace{-1.0cm}
\begin{IEEEbiography}[{\includegraphics[width=1in,height=1.25in,clip,keepaspectratio]{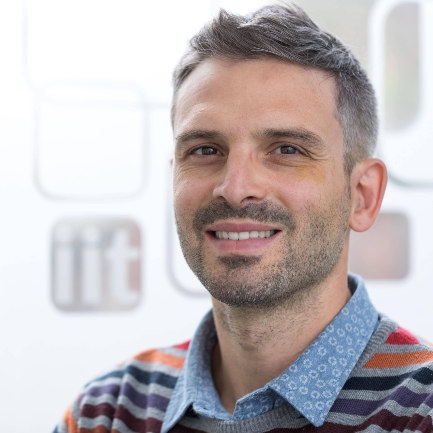}}]{Michele Focchi}
	is currently a Researcher at the DLS team in IIT. He received both the Bsc. and the Msc. in Control System Engineering from Politecnico di Milano. After gaining some R\&D experience in the industry, 
	in 2009 he joined IIT where he developed a micro-turbine for which he obtained 
	an international patent and a prize. In 2013, he got a	PhD  in robotics, 
	getting involved in the Hydraulically Actuated	Quadruped Robot (HyQ) project. 
	He initially was developing torque	controllers for locomotion purposes, subsequently
	 he moved to higher	level (whole-body) controllers and model identification. He was also
	investigating locomotion strategies that are robust to uncertainties and work 
	reliably on the real platform. Currently his research interests are focused on 
	pushing the performances of quadruped robots in traversing unstructured environments, 
	by using optimization-based planning	strategies to perform dynamic motion planning.
	 He published more than 42	papers in international journals and conferences and supervised several master and PhD thesis.
\end{IEEEbiography}
\vspace{-0.8cm}
\begin{IEEEbiography}[{\includegraphics[width=1in,height=1.25in,clip,keepaspectratio]{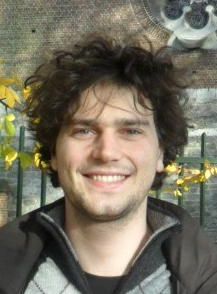}}]{Mario Zanon}
	received the Master's degree in Mechatronics from the University of Trento, and the Dipl\^{o}me d'Ing\'{e}nieur from the Ecole Centrale Paris, in 2010. After research stays at the KU Leuven, University of Bayreuth, Chalmers University, and the University of Freiburg he received the Ph.D. degree in Electrical Engineering from the KU Leuven in November 2015. He held a Post-Doc researcher position at Chalmers University until the end of 2017 and is now Assistant Professor at the IMT School for Advanced Studies Lucca. His research interests include numerical methods for optimization, economic MPC, optimal control and estimation of nonlinear dynamic systems, in particular for aerospace and automotive applications.
\end{IEEEbiography}

\begin{IEEEbiography}[{\includegraphics[width=1in,height=1.25in,clip,keepaspectratio]{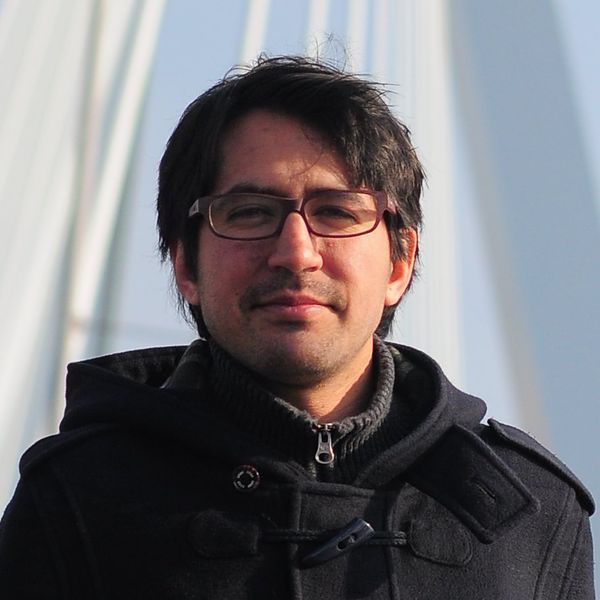}}]{OCTAVIO VILLARREAL}
	is currently a postdoctoral researcher at the Dynamic Legged System (DLS) lab of the Istituto Italiano di Tecnologia (IIT). He receive his MSc degree in Mechanical Engineering track Control Engineering from TUDelft in the Netherlands in 2016. He received his PhD from IIT and the University of Genoa on 2019 working at the DLS for his work on vision-based foothold adaptation and locomotion strategies for legged robots. His current research interests are related to control of legged robots, vision-based locomotion and foothold adaptation strategies, model predictive control for legged robots and trajectory optimization.
\end{IEEEbiography}
\vspace{-1.6cm}
\begin{IEEEbiography}[{\includegraphics[width=1in,height=1.25in,clip,keepaspectratio]{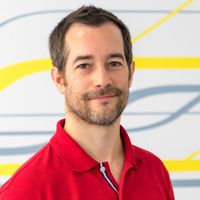}}]{Claudio Semini}
	(MSc 2005, PhD 2010) is the head of the Dynamic Legged Systems (DLS) lab at Istituto Italiano di Tecnologia (IIT) that developed a number of high-performance hydraulic robots, including HyQ, HyQ2Max, and HyQReal. He holds an MSc degree from ETH Zurich in electrical engineering and information technology. He spent 2 years in Tokyo for his research: MSc thesis at the Hirose Lab at Tokyo Tech and staff engineer at the Toshiba R\&D center in Kawasaki working on mobile service robotics. During his PhD and subsequent PostDoc at IIT, he developed the quadruped robot HyQ and worked on its control. Since 2012 he leads the DLS lab. Claudio Semini is the author and co-author of more than 100 peer-reviewed publications in international journals and conferences and he received several awards for them. He is also a co-founder of the Technical Committee on Mechanisms and Design of the IEEE-RAS Society. He is/was the coordinator/partner of several EU-, National and Industrial projects (including HyQ-REAL, INAIL Teleop, Moog@IIT joint lab, ESA-ANT, etc). His research interests include the construction and control of highly dynamic and versatile legged robots for field application in real-world operations, locomotion, hydraulic drives, and others.
\end{IEEEbiography}
\vspace{-1.6cm}
\begin{IEEEbiography}[{\includegraphics[width=1in,height=1.25in,clip,keepaspectratio]{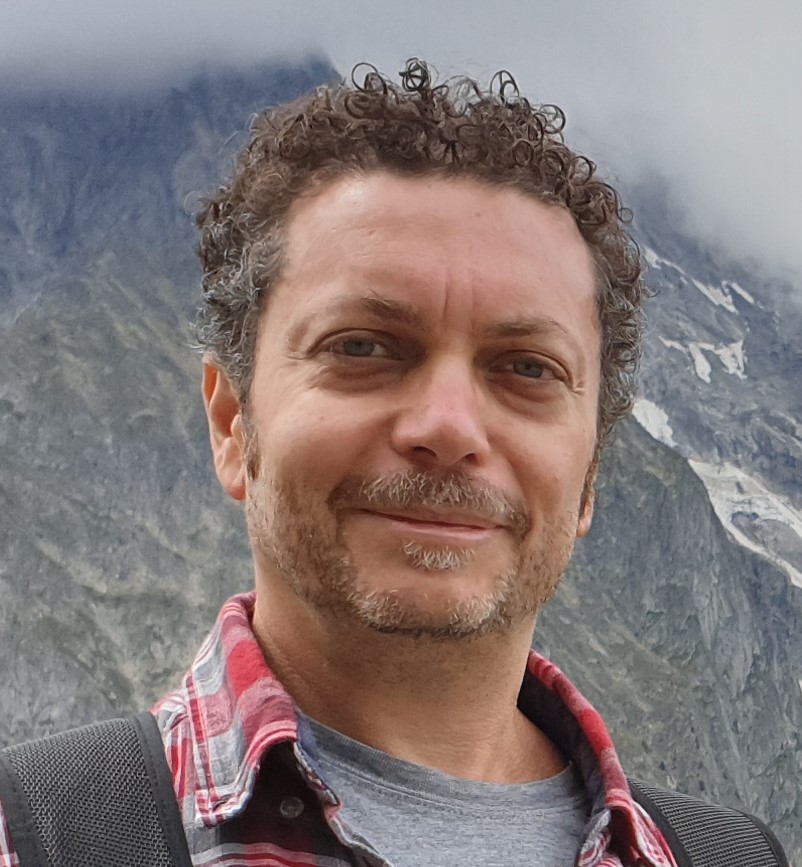}}]{Alberto Bemporad}
 received his Master's degree in Electrical Engineering in 1993 and his Ph.D. in Control Engineering in 1997 from the University of Florence, Italy. In 1996/97 he was with the Center for Robotics and Automation, Department of Systems Science \& Mathematics, Washington University, St. Louis. In 1997-1999 he held a postdoctoral position at the Automatic Control Laboratory, ETH Zurich, Switzerland, where he collaborated as a senior researcher until 2002. In 1999-2009 he was with the Department of Information Engineering of the University of Siena, Italy, becoming an Associate Professor in 2005. 
 In 2010-2011 he was with the Department of Mechanical and Structural Engineering of the University of Trento, Italy. Since 2011 he is Full Professor at the IMT School for Advanced Studies Lucca, Italy, where he served as the Director of the institute in 2012-2015. He spent visiting periods at Stanford University, University of Michigan, and Zhejiang University. In 2011 he co-founded ODYS S.r.l., a company specialized in developing model predictive control systems for industrial production. He has published more than 400 papers in the areas of model predictive control, hybrid systems, optimization, automotive control, and is the co-inventor of 16 patents. He is author or coauthor of various software packages for model predictive control design and implementation, including the Model Predictive Control Toolbox (The Mathworks, Inc.) and the Hybrid Toolbox for MATLAB. He was an Associate Editor of the IEEE Transactions on Automatic Control during 2001-2004 and Chair of the Technical Committee on Hybrid Systems of the IEEE Control Systems Society in 2002-2010. He received the IFAC High-Impact Paper Award for the 2011-14 triennial and the IEEE CSS Transition to Practice Award in 2019. He is an IEEE Fellow since 2010.
\end{IEEEbiography}
\end{document}